# Conscious Intelligent Systems

Natural Intelligence and Consciousness – A Learning System Perspective

## Part I: I X I

Preface

Designs from Nature

Armchair AI

Consciousness

Mobility

Consciousness Driven Learning Systems

Dilemma

Solution?

Mind

Presence of Mind

I

Movies in My Mind

Database, Learning, Patterns, Search - A Question of Scale

Designs of Nature

## I X I

### Preface

"This work contains hardly any original facts in regard to man; but as the conclusions at which I arrived, after drawing up a rough draft, appeared to me interesting, I thought that they might interest others… The conclusion that man is the co-descendant with other species of some ancient, lower, and extinct form is not in any degree new." - Charles Darwin – The Descent of Man

**Q**uestions about the human mind and consciousness are old questions, possibly man's oldest. What is consciousness? What is mind? Who am I? The trend of the moment is to look towards the natural sciences for help in breaking the brain-mind-consciousness lock. It may look foolhardy and unseasonable for someone at this point to take an old-fashioned logical hypothesis based look at these problems. And when such an approach originates from first principles, it can look nothing less than quixotic. However this is what we have set ourselves to do, we take here a first principles, learning system perspective to these problems, perhaps for once last time. Let us start tilting at windmills.

The brain and the computer may not be similar, but no one disagrees with the proposition that natural entities are also learning systems. One can see that if natural entities are learning systems, then such systems need to follow some kind of logical design or design path. Is it possible to unravel the design paths of natural learning systems? Here we make an effort, a humbler one sans modern tools, and see if we can derive a logical learning-system-based explanation for the rise and presence of mind, our sense of I, and its relation to consciousness.

To that aim, let us leave aside the beaten track. Rather than undertake a detailed study of natural intelligence systems, let us do a simpler thought experiment. Let us assume that nature started off with a simple learning system. Lets us then ask, in tribute to Simon's ant, what kind of learning conditions could have possibly given rise to mind and consciousness. Such a thought experiment and its mapping to the natural world and to humans does give rise to some interesting possibilities, and can allow for a more coherent and natural explanation for natural intelligence system phenomena. Let us take a brief overview.

# Conscious Intelligent Systems

Natural Intelligence and Consciousness – A Learning System Perspective

## Part I: I X I

We first posit a simple learning system and then a complex learning system in a natural environment and see how environmental constraints and enablers shape their growth paths. Here we understand the importance of archive based environmental knowledge and the need for pattern based learning. We understand the possible role and importance of reproduction and incubation to natural life. We discover how natural environments force all learning systems to abandon online learning in favor of online response. We see how this push results in the rise of offline learning mechanisms. We see how this offline movement can ironically result in better learning and how it may have even contributed to our rise and presence on this planet. We see how the presence of consciousness and intent enable natural entities with offline learning mechanisms to assume a semblance of control over their local environments.

We define consciousness in the simplest terms possible and use this definition to build a consciousness driven learning system. We discuss how mobility influences the growth of consciousness and intelligence in consciousness driven learning systems. We discuss the effects of natural environments on such consciousness driven learning systems and how it can result in the rise of the mind and our sense of I. We see how minds may have online and offline modes. We look at the presence and need of sleep from a learning system perspective. We posit reasons for the rise of man's mind, his mixed mode of mindlessness and mindfulness and his incessantly looping thought processes.

We do not attempt anything radically new in this discussion other than a slight shift in perspective, a shift that allows natural phenomena to fall more coherently and naturally in place. If the evolution of life forms is considered to have risen out of an interaction between living organisms and the environment, here we presume that the evolution of consciousness and intelligence systems must have risen through a set of interactions between learning systems and environmental factors and would have followed a similar evolutionary path. The study of evolution perhaps forms a close analogy to our approach. Intelligence theorists and scientists working on understanding mind and consciousness may find such a change in perspective and the resulting discussion meaningful and interesting.

**Pre Script**: This discussion takes the chance that consciousness, a seemingly complex entity, if of material origin and systemic, rests on simple design principles. Consciousness is both a siren and a mirage; it attracts avid enthusiasts only to let them down when they get too close. Much like an Indian God with many arms, it takes a multiplicity of meanings and shapes. Experience tells us that nature's basic mechanisms are generally simple and complex looking natural systems are generally multiple iterations or modifications over simple systems. Could consciousness be such an artifact?

We take the simplest possible approach and we base it on simple assumptions. Our attempt is to see if we can derive a probable, functional, and extensible framework for mind and consciousness design, a frame work that also lends itself to being falsifiable, an important criterion for a theory as any. The final system that we come up with could perhaps be sketched on a sheet of paper; this system is but an ant.

A framework that covers such a huge area has to be necessarily panoptic and bare boned. Such an approach however also makes the following discussion devoid of much of the existing major texts and subtexts of consciousness and learning theories, also much of biology and philosophy. While we lay out a possible design path, we avoid any discussion of its implementation in the natural domain. A sizable amount of hard evidence is just beginning to emerge in these areas and as is true of any new field they seem to be open to multiple explanations and interpretations. This forces us to ignore most of it and follow the logic of the perspective, which is at best a dangerous course for anyone trying to explain natural systems.

This discussion is therefore rather speculative, more a discussion of possibilities and does not claim exactness or correctness in detail. The well known neuroscientist VS Ramachandran once quoted Medawar's "We are not cows grazing on the pasture of knowledge" (Frontline interview, April 2006) to defend logic based non-QED style speculation. No one could disagree; the seeds of science have to first sprout in the mind before they take root on more solid ground.

(In a companion paper that forms Part II of this discussion (5) we use the results of this discussion to reexamine the idea of understanding and see how the presence of mind affects its communication and discover how minds cause language. From such a perspective, we also reexamine the validity of mankind's perennial humanoid fantasies. We find that to reach human levels, AI entities may need to be something more than the learning machines they are now.)

# Conscious Intelligent Systems
Natural Intelligence and Consciousness – A Learning System Perspective
## Part I: I X I

### Designs from Nature

"You get a lot of respect for natural biological systems. Even ants do all these functions effortlessly. It is very hard for us to imitate that and put it into our machines"  - Reinhold Behringer - quoted in a National Geographic Nov 2004 article on autonomous robotic vehicles.

Natural entities and their underlying intelligence systems have evolved through long periods of geological history in response to a variety of environmental and evolutionary challenges. Natural organisms demonstrate multiple levels of intelligence and consciousness.

The common strain through out nature and evolution is that these natural entities are exposed to recurrent survival risks and unending environmental variation. The resultant learning aims and response demands placed on a natural entity are thus different from what is generally provided to an artificial intelligence mechanism or entity, static or mobile.

Nature's primary focus is first on entity survival and then on entity comfort. Entities that need to survive in nature like environments need to demonstrate quick and appropriate environment responses. Such responses arise from the learning processes and intelligence systems within the entity. All learning processes unfortunately exhibit processing or learning delays. In natural environments such learning delays create response delays that could be inimical to the very survival of the entity.

Response speed is therefore very critical in nature like environments. Response appropriateness however arises out of good learning. Therefore, the target for any natural intelligence system should be to find the right blend of good learning and quick response, the latter being more important in the short term. The design of intelligence systems in nature like environments should contain design factors and paths that help achieve such good blending in aid of both short and long term targets.

The learning load for any entity inhabiting a natural environment comes from its system sustenance demands and environmental challenges. Natural entities generally gather sustenance based on what the environment provides, therefore environmental variation forms the bigger part of entity learning load.

Learning loads are generally not uniform in a natural environment. They vary both in time and over time. Even seemingly stable environments are not immune from such dynamic load patterns. We know that for any learning or data processing system, peak processing or learning loads can lead to system stall. In nature, such system stalls could prove fatal. Therefore, any learning or intelligence system that needs to learn and survive in a natural environment has to take cognizance of dynamic learning loads and manage to keep it within control to avoid/reduce fatal risks.

Natural intelligence systems seem to have learnt to overcome such dangerous dynamic learning overloads and other natural load variants. Three of them seem important to us from a design point of view, stall control, load balancing, and proactive behavior. We will discuss these solutions where appropriate.

In stall control, the factors' leading to system stall are avoided and includes an emergency response system. In load balancing, natural systems take advantage of low dynamic load periods by shifting peak learning demands to such periods. With proactive behavior, the natural entity seeks and acts to reduce learning loads by reducing either its exposure to the environment or the environmental variation it needs to face, to the limits of its learning system capacities. While we will discuss how these solutions arise, natural scientists are best posited to show and understand how natural intelligence systems demonstrate such learning load reduction behavior.

Such solutions imply that natural learning systems are not environment reactive systems or mere environment punch bags buffeted by environmental demands; they are autonomous intentional systems that are conscious of their needs and act to satisfy them while keeping their exposure to environmental dangers under check. Such intentional, separatist activity demands a basic sense of distinction or awareness about themselves - an awareness of themselves as entities separate from

# Conscious Intelligent Systems

Natural Intelligence and Consciousness – A Learning System Perspective

## Part I: I X I

their environments and interacting with it for sustenance and protection. For this discussion we assume that such awareness and directed activity derives from the presence of life.

It is the awareness of such differentiation that we call consciousness. Such differentiation from the environment and the need for efficient interaction with it lead directly to the need for learning systems. This implies that life necessitates natural consciousness, which in turn necessitates natural learning systems. Vice versa, sans consciousness, the need for a learning system does not arise. What follows in this discussion is an elaboration of this simple understanding. In doing so, we elucidate how a consciousness driven learning system can arise and more importantly how natural constraints and enablers act to orient and influence natural learning system growth paths.

For such a purpose we create a simple, functional definition of consciousness based on a simple assumption about life. The main challenge for such a definition of consciousness and the resulting consciousness-based learning system would be to explain the phenomenon of human self-consciousness. We will see how such self-consciousness could arise from a simple conjunction of learning system evolution and human evolutionary history.

To keep the discussion short and simple, we are forced to take a snapshot view of the growth of conscious learning systems. Readers can read and extrapolate as to how natural settings, learning load, and consciousness interact to coerce and influence natural learning system design and how such interaction has resulted in animals and man. Given the recent mushrooming of evidence from natural science studies the real test of our system lies in the degree of fit it can demonstrate with such emerging evidence.

**Armchair AI**

We rush through the basics; professionals can perhaps skip this section and start directly with the next section, which discusses consciousness and then work back if necessary. However in this section arise first principles, design insights, and features that set the ground for later discussions.

As mentioned earlier, we posit a simple learning system in a natural environment and help it grow to deal with the basic demands of the natural environment. Here you will notice that we do take certain capacities and facilities as granted, we would however not fail to elaborate on such assumptions as we move on. The scenarios discussed here on are hypothetical but they help cover the ground faster. The discussion is deliberately kept simple so that only the major points show up and it proves a quick read.

A simple learning system is a system that consists of an environment sensor and a processing mechanism that can learn in a given environment. We will consider a natural environment as being characterized by dynamic learning load patterns, time constrained response demands, and pattern like environmental repetition.

In such dynamic and time constrained environments a simple learning system, whatever be its internal design, is susceptible to system freezes, particularly under a combination of high dynamic processing loads and time constrained response demands. In a natural environment, where we posit our system to be, the learning needs of the system arise from the environment itself and the interplay of its inhabitants. Repetition is a general characteristic of all environments, natural or artificial; therefore if we can add a learning archive to our system, then we can store the successful results of learning and reuse it when the environmental challenge returns.

Such an archive can help reduce repeated learning demand and the system can offer at least partial solutions from its archives to problems that show repetition. We can see that old solutions need not always fit in well to an environmental challenge even when it shows repetition, given that natural environments show repetitive but chaos like behavior. Chaos-like behavior is characterized by bounded, but infinitely repetitive, and therefore infinitely variant system behavior. A learning system in such an environment has to learn to modify its existing solutions on the run to deal with such small but non-foreseeable variations. Run time adaptation makes sense in nature like environments.

# Conscious Intelligent Systems
Natural Intelligence and Consciousness – A Learning System Perspective
## Part I: I X I

This does imply that there are no perfect solutions; we have a bunch of solutions that demonstrate varying degrees of fit to a bunch of environmental challenges. Such solution reuse helps reduce the high dynamic processing power demands that result when solution processing has to proceed from scratch. Such solution clumping also reduces data storage requirements and long term processing requirements. The power of an archival solution strategy comes to the fore during an emergency when even a partial solution can be lifesaving to an entity that could otherwise freeze as a result of highly dynamic learning demands coupled with extreme time constraints.

Emergencies are also intrinsic to natural environments and are actually a reflection of both the environment's challenge and the entity's preparedness to meet it. In an emergency, the usual time available for processing is refused and in the absence of a prior solution, entity survival can depend on mere chance and environmental benevolence. This is a major possibly fatal risk, no conscious learning system interested in its survival will be able to take, and unless learning systems can learn from such tight situations, they will be as ill prepared to meet a future emergency as they were in the first instance.

This creates a dilemma for the learning system; on one hand, extreme time constraints do not allow it to learn, and on the other hand, without learning from an emergency, it cannot better its response or reduce its susceptibility to risk on a future occasion. On reflection this applies to all time constrained learning processes. How do we design a learning system that can learn under such time constraints?

The possible way out is to store emergency data in an archive and look for a time when we can reprocess or learn from this data, such times are available in natural environments as rest time. It could be the night, or in general any time when environmental challenges are minimal and do not stress the entity.

Notice that even with our earlier strategy of learnt response archival, a fair amount of data storage becomes necessary. This is because any archive-based response process depends on environment challenge cognition. Such problem cognition can only happen if similar data exist in the archive, which implies some amount of data storage. Such storage of data however means that the learning archive grows really big and fast, which increases search time. Large archives that need searching however imply delayed cognition, resulting in delayed responses and increased risk. So archive search processes and cognitive processes need to be very fast for archival solutions to work. We will deal with this problem at later points in the discussion.

The presence of the archive and streaming data means that data storage and archive management become important activities of the system. When will the system find the time to do all this, without debilitating the entity's normal environmental response? On the other hand without such management processes, the presence of the archive will really become a drag on the system. A real time online learning system will demonstrate better learning in the short term.

We will however retain the archive-based design since its long-term advantages outweigh the disadvantages. In nature, where time for online dynamic processing is frequently limited by external factors, the first advantage of an archive-based system with a built in cognition process is that a prompt if not exact response, is guaranteed. This advantage can be a life and death differentiator for the learning system in a natural environment.

Archival solutions also bring forward another possibility, that of learning retention, and by extension, the mechanism of reproduction. Learning from scratch is a costly and risky endeavor for any learning system that aims to survive in nature like environments. It is obvious that retaining the results of learning and reproducing them can help successor or descendant systems start with an advantage, particularly with respect to an environment of medium to long-term stability, where the environment is itself the cause of learning loads.

As long as environments remain stable, the chances of survival of such descendant systems remain high. The presence of the knowledge bank means that systems do not need to worry about normal environmental challenges. The use of reproductive strategies and incremental learning mean that with time, the data bank grows higher and richer. On the whole processing power requirements in the long term reduce. This also means that entities can use their existing processing power to deal with new situations and add such learning to the knowledge bank, so that after multiple generations, the species as a whole can deal with its environment on a more comfortable basis.

Such a quickly reconfigurable iterative solution process that exists on partial solution bases is good for chaos like repetitive environments which themselves show slow variation, because it allows descendants to keep pace with the environment, without being locked on to hard bound solutions. Linear zero to hero learning is not a viable option for natural entities, such an option demands very high real time processing power, which may not be practically possible. (An ant with the brain of a man?)

Our simple learning system with the addition of archives, search processes, and environment cognition is better termed a simple intelligent system. The primary reason is that even if the system does not learn continuously, it has the choice of either learning/responding online or environment cognition/archival solution reuse. This enhances both entity comfort and entity survival possibilities.

Our simple intelligent system with its archive-based design can exhibit a small measure of comfort in environments of low variance. However, exposing it to higher variance can bring about system stall, which as we know can threaten survival.

High variance environments bring in a data inrush so high that the processing mechanisms will find it difficult to keep pace with it. The major problem is in the architecture we normally provide for learning systems. The usual process and forward architecture means that data coming in from a sensor, goes into a small data buffer, from where it moves in to the processing mechanism for solution generation and delivery. When environments speed up, data will tend to pile up at the processing end, where the inevitable processing delay means that the data that cannot be accommodated in the data buffer is lost.

Even otherwise, processing delays during such an environment speed up will mean that the systems responses do not keep pace with environmental response demands and therefore the system as a whole goes out of tune with its environment. Under very high data deluges, the system may even stall; there is a distinct call for a reboot. In a natural environment, any possibility of system stall needs to be avoided because once a system stalls, even an emergency response like hiding can become impossible, the system in effect is forced to depend on chance and environmental benevolence. Real life allows no reboots.

We will modify the system so as to avoid the possibility of system stall and the corresponding data loss during an emergency. We will attempt to store the maximum data possible even through the emergency so as to make it available for rest time processing. Such processing can at least result in improved cognition if not appropriate action when the emergency conditions repeat.

For this purpose, we first isolate the data queue and the processing queue. We will stream incoming sensor data directly into a separate sensor memory bank bypassing the processing queue. All sensor data thus flows into the memory bank and is stored, from where it can be recalled for later processing. Next we give the processor the choice of withholding processing or doing selective processing. To make a choice, the processing unit will poll the data queue for incoming data speeds. When data speeds are abnormal, it will alert an emergency control system and hand over control to it. Once the emergency is past, normal processing can restart.

When data speeds are normal, even as data flows into the memory banks, the processing mechanisms will dip in to this data stream and process the data. In the process and forward architecture, all data that arrived in the queue had to be necessarily processed; here we avoid such a necessity. This not only reduces processing time wasted in processing non-critical data, but also reduces processing delays inherent to such wasteful processing.

The ability to do selective processing is a great help in cognitive architectures, since it is possible for the system to sieve the data stream and pick only the requisite data to process. Such data that already have solutions in the archive can be diverted into response reuse or response adaptation modes that demand very little processing power compared to complete online real time data processing. This means that more processing power and time is available for the fresher data and this improves the learning of the system in the long run. The other advantage of selective processing is the possibility for intentional processing, where the conscious processing mechanism based on its priorities can choose the data that it wishes to process.

Both these modifications allow the system to process the more important data even as an emergency is in progress, which means that the system does the maximum possible processing before handing over control to an emergency response process. Even if its response fails, all possible data is stored and can be reprocessed at a later time so that when such emergency conditions repeat, it can be recognized and at least an evasive action can be taken.

These modifications also allow for saner processing during normal conditions, the pressure for knee jerk processing is reduced and as a result the system as a whole is in human terms less anxious. If the system can learn during its rest time from the stored emergency data and other data that lies unprocessed in the archive, then it is clear that our system can, in time and over learning cycles, improve its responses to environmental challenges.

In hindsight, our supposition of a system with a single sensor is not practical in natural environments. Natural entities are generally equipped with multiple sensors. However when we have our simple intelligent system acquire multiple sensors, then it no longer remains a simple intelligent system, it necessarily becomes a complex intelligent system.

Even without going into the detailed system architecture, it becomes clear that the resultant multiple sensor based system needs an architectural revamp and addition of processing power. In line with our earlier design, each sensor also demands a separate streaming archive and archive management becomes a major and time-consuming task. It is clear that such time consuming processes cannot be done in real time without debilitating the systems online response. For the moment however, for the purpose of moving forward, we will assume that such a multi sensor, sensory bank based, archival, cognitive, offline learning architecture exists. You were forewarned; we are rushing through the subject.

It can be seen that given a natural environment, such a multi sensor system, despite our system stall workaround solutions, can still fail. On one hand, instantaneous learning loads have not only multiplied due to the presence of multiple sensors, but the need for centralized and combinational data processing means that cumulative processing demands are very high, therefore learning delays can increase.

On the other hand, it is to be noticed that natural time constrained response demands are blind to such increases in learning system power or system complexity. Nature is not bothered about the system complexity or learning quality of its inhabitants, it demands that learning systems in natural environments display appropriate timely responses to environmental challenges.

Given the possibility that a static multi sensor complex intelligent system can fail in a natural environment, the chances of failure for a mobile entity equipped with such a system are simply higher. Mobility implies constantly changing learning loads and a complex multi sensor intelligent system in a mobile entity is certainly doomed to failure. One can deem increasingly higher processing power for such entities, until one remembers that in nature, processing power provision and availability are subject to morphological and evolutionary constraints.

We can see that the system has a much better chance of survival if it can do away with real time learning and can concentrate on search and response processes from its archive. Learning though necessary can come later; prompt responses and survival are primary requirements. However, with a multi sensor complex intelligent system, even such prompt responses cannot be guaranteed. This is because the presence of multiple sensors multiplies not only data flow but also archive sizes, simple linear searching through a multi column database can take time and demand multiple iterations. Parallel searches can reduce the time and iteration required for such searches, but still search, cognition and response delivery are going to be delayed affairs. With both search and response impaired and online dynamic learning being difficult, the risks for a multi sensor complex system in a nature like environment are simply too great.

A possible savior that can enable faster cognition in multi sensor arrangements arrives in the form of patterns. Patterns, an idea familiar to AI people, arise from and are endemic to stable environments. A pattern can mean a certain set of environmental conditions or can indicate a sequential environmental condition chain. Patterns can generally be recognized from the presence of leading incoming data or pattern markers. The presence of pattern markers can ease the pattern cognition and response planning process. Pattern cognition implies easier cognizance of the incoming environment.

If such pattern identification and pattern cognition are possible (we skip the details of the pattern identification and cognition process) then the entity's environment response can improve over time and dramatically. Pattern cognition allows the entity some breathing space because the entity gains a fair amount of pre-knowledge of the possible environment path. Such pre-knowledge can be a great comfort to the time harassed entity as it can plan or choose a possible response path to the incoming environment.

If identified patterns and their solutions can be put in a separate archive then searching this pattern archive is easier; since in an entire lifetime, the number of possible patterns an entity will encounter are infinitesimal when compared to the number of probable combinations a multi-sensor intersection of the environment can create. Given an environment, natural or artificial, its possible patterns are limited in number; this makes pattern archives small and close to linear, thus making search not only easier but also faster. The presence of pattern archives can drastically cut short search time and improve environment cognition/response time.

Pattern identification is however not an easy process, moreover it is an extremely time consuming and resource consuming process. A good knowledge of patterns can arise out of archive study. Fortunately such an archive study process does not demand real time data processing; it is actually better suited for offline processing. Archive management, we did discuss earlier is best done offline to avoid debilitating the entity's response to the environment, now we see that the process of archive study and pattern identification is also best done offline.

Such offline movement however necessitates rest periods when environmental challenges and demands are low. Fortunately such periods and facilities are available and routine in natural environments. Rest periods are good times for offline activities like pattern identification, archive-management, and offline learning. Rest periods are therefore a necessary artifact for intelligent archive based entities in such environments. When a rat hides in its hole, it is a nice time for such data processing activity. It can be guessed that when environmental challenges rise and learning demands increase, short term or long term, the intelligent entity will be forced to find rest-enabled times, areas and environments, within the limitations of the environment.

This however means that unlike usual AI entities, natural learning entities cannot switch off power once their online response demands are over. They will have to stay powered, so that they can undertake offline activities like these during rest periods. Switching off power translates to death for most natural entities, there is also increasing evidence that offline time could be learning time for nature's entities. While the presence of archival processes save them from known environmental challenges, such offline processes can help save them from newer non-processed environmental challenges.

In nature like scenarios, where processing power straddles a broad range, where offline periods may themselves be low, and lifetimes are short, pattern identification and learning may not always be possible even within an organism's lifetime. We can see that these are resource intensive costly processes, so it makes sense to retain the lessons of learning across generations, rather than have each generation acquire it from scratch. This demand for conservation of learning necessitates a suitable knowledge retention and reproduction mechanism.

With patterns, the call for reproductive mechanisms, a call that did arise earlier from the presence of a learning archive, becomes clearer and cleaner. In hindsight, simple archive reproduction is inherently messy, and does have many disadvantages. A lot of this messiness can be reduced with pattern archive reproduction, but an even better option is to bypass the archival pattern data and pass pattern templates.

Such a pattern template reproduction option, as computer scientists can understand, would not only reduce archival carrying loads, but also reduce the demand for exact archival matching. This does demand the use of approximation during cognition, on the other hand such templates are easy to standardize, carry, install and use. The greater disadvantage of template reproduction is the demand for template fill-in before these patterns can be used online.

Translated to nature, such a fill in demand means that a call for incubation arises. Incubation entails both incubation time and incubatory protection processes. This means that parental or environmental protection should be available to the emergent entity. Incubation time and experiences allow these barebones templates to be filled in before the entity is called to directly war with the environment. More complex and more intricate the templates, higher are the need for incubation experiences and incubation time.

When learning loads increase and incubatory facilities are available, limits on processing power may demand a phased learning load rollout during the incubation period, so that learning acquisition can stay within the processing power of the underlying intelligent system. This in turn means that the entity's ability to engage the environment from birth is debilitated and a growth sequence for the new born entity, in consonance with its processing powers is required.

This also means that with good incubation facilities and longer lifetimes, entities with increasing learning loads can be rolled out. One can see that the final product, mechanism and roll out of an incubation based, archive based natural intelligence solution is a conjunction of multiple factors. Processing power limitations and the resulting incubatory strategies are factors that turn simple (gene based?) reproductive blueprints into recipes, each with its distinctive growth path. It is clear that the choice of reproduction as a learning retention mechanism will have and has had great implications in the design, structure, and function of natural entities, entity life styles and lifetimes in the natural kingdom.

In hindsight it is obvious that such a reproduction strategy is only possible with an archive-based design. The presence of reproduced archives and archival solutions can instantaneously reduce the starting learning demand and lifetime learning demands of successor generations. Incubation helps deliver a product that is ready to perform in an environment and which can continue learning from where the last generation left off. In effect, reproduction and incubation allow species to run relay races covering long geographical periods with little processing power. If a reproduction based archive/pattern passing process were not available, it can be foreseen that all natural organisms would have to start learning from scratch; the demand for processing power would be very high and life would have found it extremely difficult to pass even the bacterial stage.

Let us get back to patterns, because patterns provide us with more interesting possibilities. Why do we harp on patterns so? It is because of the understanding that even the simplest natural mobile learning entities are pattern-based entities. Every process in life and evolution is tied to patterns and the abilities that pattern bases provide to natural learning organisms. Read on and you will find this a statement you will fall naturally in agreement with.

We did see that pattern knowledge allows us to predict environments from pattern markers. Iterative archival study and pattern correlation can let the system improve its knowledge of pattern markers, and choose markers that can predict incoming environmental conditions as early as is possible. We can see that for an entity that is equipped with a facility for such environmental prediction, the environment is no longer a perennially variant, possibly malicious master; it almost becomes a predictable sequence of actions and actors. The possibility that a pattern-based entity need not wait for the environment to completely expose itself, before it can plan a response gives entities the time and comfort of predetermining its response chain, a response chain that can allow it to fall seamlessly in step with an incoming environment.

# Conscious Intelligent Systems
Natural Intelligence and Consciousness – A Learning System Perspective
## Part I: I X I

The availability of this time gap between environment cognition and environment response also gives rise to some rather interesting possibilities. For instance given that the environment is open to prediction, given that there is a time gap available between environment cognizance and actual time of response, given that environment feedback is (also) dependent on entity response activity, is it possible for an entity to preplan and present an action that will influence the environment, even if in a limited sense, to provide a beneficial feedback response? Or at least reduce the entity's susceptibility to an adverse response?

In effect, rather than stay put with simple environment prediction and passive response play out, is it possible for a pattern based entity to actively engage the environment so as to maximize possible environmental benefits and minimize environmental dangers? Can the entity ditch passive environment responsive behavior in favor of environment ahead proactive behavior?

For such proactive behavior, it is clear that the entity can no longer be a pure environment responsive entity; it should be an intentional entity that seeks to maximize its environmental comfort. We will see in the next section how life and consciousness embed all natural entities with intent. Such intent can allow these entities to demonstrate dynamic load management behavior and environment-ahead proactive behavior.

Even without going into a detailed discussion of intentional environment ahead proactive behavior, we can see that the very possibility of such environment prediction and environment management by an intelligent entity tends to turn our general perception of the objectives of environment based AI learning systems on its head. The general target for AI entities operating in artificial and natural environments is good environment responsiveness. We are yet to consider the possibility of active environment management; an activity that nature's entities engage naturally in.

Such environment prediction and environment preemptive action allows entities to gain a semblance of control over their local environments and can make life easier for the entity. This implies that for natural entities that have occupied natural environments over generations, the real target is not environment responsiveness. Such base targets have long been overshot through generations of learning; the real target of most natural entities is environment comfort.

The idea behind environment comfort is not the idea of comfort per se; it arises as an unintended side benefit of consciousness based learning system design. While still being dependent on the environment to a large extent, intentional pattern based systems can and will, like an intelligent servant (Jeeves?) take advantage of the master when and where possible, even arm twisting him like a child sometimes arm twists its parents.

However not all is fine in the pattern based world. We can see that environment stability gives rise to pattern based systems; higher the stability and longer the period of stability, more intricate and detailed will be the pattern database. The possibility of response optimization will mean that as a species, these natural entities will over time, learning cycles, and generations increasingly dovetail themselves to their environments.

Such dovetailing is not always beneficial to the entity or its species. Dovetailing increases the detail and complexity of patterns and increasingly fine grains the entity's responses to the environment. When environments change drastically, as they sometimes do, simple patterned systems can relearn and reorient themselves to most changes quickly, advanced patterned systems will find that difficult to do.

When environments change drastically, existing patterns may need to be abandoned, new patterns will need to be identified, and newly identified patterns need to be written back into the genetic database, all under time pressure and in very short time spans. In effect the learning correction backlog becomes very high; relearning disturbs the normal life cycle of the entity thus exposing it to increasing stress. Entities that depended on archival strategies and lived in comparative comfort, with minimal processing power will find such inflated learning demands difficult to manage. Hardwiring mechanisms, if they exist, can worsen the scenario; all instinctual responses can become suspect in the new environment and this can spell doom for the organism and its species.

# Conscious Intelligent Systems

Natural Intelligence and Consciousness – A Learning System Perspective

## Part I: I X I

Therefore, it can be seen that when the environment changes faster than a pattern-based entity can learn from it, a separate reason for species extinction is not required. The tendency towards extinction or survival is dependent on the speed of the environment change envelope, the faster it moves, higher the chances of extinction. Entities that survive such drastic environmental changes are usually the beneficiaries of environmental or ecological niches. Entities that live in environments that are always at the edge of change also have better chances of survival since their processing power is not only perennially active but also slightly higher than that of environmentally comfortable organisms.

Thus we see that Darwinian natural selection scenarios can arise even for non-morphological reasons. While their morphologies may permit many natural entities to survive in environmentally modified environments, their learning systems may find it difficult to keep pace with new learning demands; the entity and its species can find their environmental responses increasingly out of tune with environmental challenges and are thus wiped out.

Over geological history, science tells us that there have been multiple periods of drastic environmental change that have destroyed millions of species. Given such repeated destruction, it can be predicted that a demand for quicker pattern rewriting abilities or quicker relearning abilities and higher processing power will rise. We will discuss in a later section, a learning rewrite mechanism that probably reduced the demand for such extinction and reduced the speed of evolution.

In our enthusiasm for patterns and their possibilities we are yet to discuss how an offline learning process may arise. In our next section we will see that the creation or implementation of such an offline learning mechanism is not as simple as pattern identification or archive management.

Given our overall experience with AI design, we can however see that real time learning can become progressively difficult in nature like environments. On one hand learning demands progressively rise because of environmental/ecological pressures, predator prey relationships and natural selection pressures and on the other hand, response demands become keener and show little regard for system complexities. Designing a real time online learning process for such a demanding environment is a challenging task, not impossible, but improbable. Offline learning is much simpler. Before we move on to such a discussion of such a system, a quick recap of our salient points on our natural learning system discussion would be:

- The presence of life differentiates entities from their environments. Consciousness arises out of such differentiation; therefore life necessitates natural consciousness, which in turn necessitates natural learning systems.

- Archive-based learning systems make sense in nature like environments. Nature like environments generally show chaos like repeated activity and provide varying time constraints to entity responses. Such time constraints can debilitate online real time learning ability and force it to be moved offline, archive based solution reuse avoids repeated learning and also offers a better chance of survival. Such solution reuse however depends on quick environment cognition, archival response look up, and response reuse, which in turn necessitate environment data storage and quick search capabilities. An emergency response system is essential for all mobile entities in natural environments.

- Processing architectures of systems that inhabit nature like variable learning demand environments will benefit if the data queue is separated from the processing queue, this reduces chances of data clog and subsequent clog induced system stall. A sensor memory bank DMA type architecture helps isolate data queue from processing queue, helps stores all environment data including emergency data, reduces data clog, allows for offline or postponed processing, and makes archive based environment cognition possible, which directly reduces online learning demands. This frees us the processing portion from having to process all data. This option also allows processing systems to choose and process data. This makes intentional processing of data possible.

- Natural environments can force even simple learning systems to concentrate on online archive based responses and move learning offline. In such an environment, multi sensor equipped intelligent systems would find online learning and response still difficult. Even archive search, cognition, and response can also become difficult because of explosive search space expansion. Such a multi sensor intelligent system can survive and benefit if it can use rest time or periods of low environment challenges to review its database, identify patterns from it, and create responses for such patterns. Pattern identification eases environment cognition and speeds up environment response in repetitive environments. Pattern identification, pattern based cognition, and pattern exploitation can enhance survival chances of time-stressed complex multi sensor based intelligent systems, therefore all multi sensor complex systems would find it advantageous to morph into offline based pattern identification systems. Most natural systems are pattern-based systems.

- Pattern based systems can exhibit comfort in stable natural environments. This is because pattern knowledge and pattern marker based cognition can help these systems predict incoming environments and therefore give them time to choose their response, which is a great relief in time constrained environments.

- However pattern identification is also a slow and resource intensive process and may not be complete within an organism's lifetime. The cost of such learning is therefore high; therefore there arises a demand for a knowledge retention mechanism across generations. In the presence of an archive, the mechanism of natural reproduction can help satisfy such a demand. Pattern template reproduction helps decrease pattern carrying loads but necessitates incubation. Good incubation strategies will allow even entitles with very high archival and learning loads to be rolled out successfully. The presence of archival lessons and incubation allow newborn natural entities to blend in quickly and seamlessly to a known environment before it needs to take on new learning loads. This archive based reproductive strategy reduces learning loads and enables even entities with low processing power to demonstrate increasingly dovetailed interactions with their environments over multiple generations.

- The presence of intent can enable such pattern-based systems to actively engage the environment rather than react passively to it, such proactive engagement can help it maximize environmental benefits and reduce environmental risks.

- When environments change quickly, pattern based entities can find it difficult to survive, because much of the pattern based knowledge needs to be rewritten and rather quickly. Unfortunately this pattern based, archive based, reproduction based design, which allowed them comfy lives with little processing power, will be found wanting in the face of massive learning loads. Quick environment changes can lead to species extinction, faster the change, higher the learning load, therefore higher dovetailing to an environment increases the species' susceptibility to extinction. Pattern based architectures which emerge from periods of environmental stability are suitable only for stable or slowly variant environments.

Our first point is that the rise or generation of any learning system is dependent on the presence of life and consciousness. So what is consciousness?

**Consciousness**

- For readers starting to read from this point, kindly read the synopsis of the earlier section in the last paragraphs to get an idea of what we have discussed until now

- At this point, informed readers have a real hard task at hand. For the course of this discussion, the author wants them to drop all notions and ideas, preconceived or received, of consciousness and related topics, so that we can proceed from a simpler perspective on consciousness.

- We start with an extremely simple and functional definition of consciousness and watch as it morphs from being the base of all learning systems to becoming its controller, directing what the entity is to learn and when and further. The author requests you to follow the logic and promises you that you will be surprised at what emerges and what the implications are. However please remember that we are only laying possible design paths and this discussion is not intended to be a final description of consciousness or its implementation in natural systems, we merely attempt to show how human like binary consciousness can arise out of a few simple design choices.

First we assume that it is inherent to life that any system that is alive tries to sustain itself so that its continuity can be ensured. While we cannot explain why this is so, we all know that it is so, we also know that we cannot explain this phenomenon based on current scientific knowledge. This is a base assumption and one of the two assumptions we use in this entire discussion.

# Conscious Intelligent Systems
Natural Intelligence and Consciousness – A Learning System Perspective
## Part I: I X I

Once life is available, then the desire for sustenance implies not only inputs to remain alive but also the protection of what is alive. For any system to sustain and protect itself, it needs to be first aware of what is to be protected. This awareness is given by consciousness.

Consciousness simply means awareness. It is logical that in any entity, awareness can arise only through its sensors. The entity's sensors delineate the entity environment boundary and all self and environmental knowledge arise from the sensory boundary. We use the term simple consciousness to indicate the system's sensory boundary image. Since environments vary but sensory boundaries remain the same, simple consciousness indicates the sensory map of the organism.

When we say sensory boundary image, we do not declare any perceptional or cognitional differences, we mean that the sensor generates sensory data when in contact with the environment. For instance, heat in an environment will map to thermal sensory data in a thermal sensor, as simple as that. The question of time differences between individual perception and base system perception and the accompanying debate are non relevant at the present stage. We will assume that all sensory data passes via suitably modeled sensory interfaces to the sensory banks and then this data will be picked up by the processing system we modeled earlier.

It is obvious that without the sensors and the resultant sensory boundary image, a rock, and a living thing are no different. It will become more obvious when you consider that any data for an intelligent system in an entity arises from the entity-environment sensory interface, without this interface and the subsequent data generation, the need for an intelligent system simply does not arise.

Learning arises and is required to make sense of and make shortcuts and rules out of this sensor data arising out of the entity environment interface (please see the section on understanding for a discussion on such learning); therefore any learning in any intelligent entity including man is always with respect to the system boundaries and on the data that arise out of such boundaries. There exists no learning demand outside of it. This can be debated; we will come back to it when we discuss understanding in a forthcoming section.

The natural question arises as to who is it that is conscious. This could be the wrong question to ask; from our definition of simple consciousness, we understand that given a sensory boundary there is no individual entity that is aware, awareness is the sum total of the effect of the presence of the system sensory boundaries.

The next natural question is that how is it that we humans are aware that we are aware. To answer that question we will have to travel a little further in this discussion where we encounter an overlying mirage like entity that is aware of and is dependent on the presence of the basic entity. Humans feel this as the involution (**def:** the act of enfolding something) of consciousness and call it self-consciousness; we will discuss how self-consciousness can arise out of such an involution. Let us keep that point aside for a moment and get back to simple consciousness.

Earlier we assumed that life chooses to sustain and protect itself. Any living system that is aware of its sensory boundaries can choose to sustain and protect itself. Since the generation of the sensory image and its sustenance and protection demands are concomitant, perhaps connatural to life, we expand our definition of consciousness to include these two activities.

The right definition of consciousness is thus system sensory image + system sustenance activity + system protective activity. This redefinition is truer to the spirit of the word than to its literal meaning. All natural life therefore needs to be conscious in the spirit of this new definition.

Our definition of simple consciousness makes it clear than an entity's simple consciousness will be limited to, and by, the coverage and abilities of its sensors. The immediate urge is to splurge on sensors; however a cost benefit analysis would dissuade us from that. In nature the spread and coverage of its sensors seem to be dependent on the cost of protection and protectability of its subsystems or components.

For instance, consider simple static natural entities, here the scope of protection is very low. Even when environments buffet them, stasis prevents these entities from taking effective protective action. The cost of protecting such a system using sensors and other artifacts will be very high. Therefore in such cases, mechanisms of easy system or component replacement have evolved and such systems are tuned for easy system or component replacement in case of damage. In their case, the sensory boundaries alter quickly and have to be remapped soon after system replacement is complete. Such remapping however becomes more difficult as complexities rise. This balance of system replacement vs. consciousness-based protection seems to be characteristic to life and is not limited to simple static entities alone.

Even within complex systems like animals and man, the system replacement vs. protection rule works at certain component levels. The idea is that if component complexity is lesser and replacement rather than protection cheaper, the need for protection of that entity, component, or sub system does not arise. For example, many of our cells are said to undergo replacement every moment and we are non the worse for it or even aware of it. However, let a small thorn nick our skin and the protective measures we take to avoid such damage are disproportional to its importance. Because of this, we can drink alcohol and damage our liver and our consciousness cares two hoots about it.

We can also see that the level and spread of consciousness can vary, in terms of the sensory image detail, system sustenance demands, and system protection demands. The sensory spread and coverage and system complexity determine to a large extent, the level of consciousness of the entity. More complex the system higher would be its protection and sustenance demands. Since one of the jobs of consciousness is the protection of the sensory boundaries, the simplest indicator of the level and spread of consciousness in a natural entity is the amount of fear and the locations it favors.

Fear/pain or similar system protective activity arises to protect the system bounds and ensure the continuity of the system's sensory image or consciousness. The other indicators for the level of consciousness are the number, type, granularity, complexity, and coverage of the sensors. The degree of centralization of the sensors, the sensory wiring, and the actuators is also a good indicator of the level of consciousness. These factors are also proportional to system replacement costs and are good indicators of system complexity.

If consciousness implies system image, system sustenance and protection, then it becomes obvious that no autonomous intelligent system can exist without first being conscious of its own sensory image and the resulting sustenance and system protection demands that emerge as a result of such an entity interacting with an environment. Any autonomous intelligent entity, natural (or artificial), perforce needs to be conscious. There seems to be no way out.

We also see that consciousness rises as a result of rising system complexity where system protection makes more sense that system replacement. System complexities tend to rise when the environment begins to demonstrate stability. We did see a parallel with patterns too. We can see how critical environmental stability has been to the rise of intelligence, consciousness, and system complexity, the comparative stability of the last sixty odd million years have helped bring us humans here. We can now link up this idea of consciousness as a collage of system image, sustenance, and protection with our earlier ideas of natural learning systems to create a consciousness driven learning system.

**Mobility**

Mobility needs good sensors; the presence of multiple sensors can help provide a better picture of the environment. Mobility also demands actuator centralization and processing. The sensors also need to be coordinated to actuation. The presence of mobility therefore leads to a cumulative demand for centralization of the sensory wiring system, the actuator wiring system, and processing. The presence of multiple sensors in a mobile entity forces it to rely on pattern study and offline data processing. The sustenance demands of mobile entities are also quite complex and varied.

# Conscious Intelligent Systems
Natural Intelligence and Consciousness – A Learning System Perspective
## Part I: I X I

In toto, all these factors that arise from being mobile add to increased processing loads and complex / centralized processing architectures. It is clear that in comparison to that of a static entity, the consciousness and intelligence system levels of a mobile entity have to be necessarily very high, the increased consciousness arising out the presence of multiple sensors and intelligence arising out of complex processing and centralized architectures.

When an entity gains mobility, each of its steps brings forth the risk of sensory boundary damage, a factor that was not present for static entities. We know that the cost of repair/replacement for mobile system components is very high due to the centralization of the sensory, processing and output architecture. This means that the demand for sensory boundary protection is immanent and permanent.

Our definition of consciousness says that the job of system protection is a function of consciousness. This implies that the consciousness mechanisms of a mobile entity should forego the typical role of a sentry that it assumes in a static system and should be transformed into a perennial patrol party, actively scouting the system boundaries for any presumed or actual danger. The mobile entity clearly needs a more active and higher level of consciousness than the static entity.

Since uncontrolled mobility can increase the risk of system damage, it is clear that such uncontrolled mobility is to be avoided, and that in the interests of system protection, mobility control be coordinated by the same system that undertakes to protect it. This means that each step and activity the mobile entity undertakes should not only be conscious but intentional. Since both consciousness and intent arise out of the presence of the consciousness mechanisms, this implies that the control of mobility should rest with the consciousness mechanisms. Consciousness in the mobile entity is therefore an active hardworking artifact doing both system protection and mobility control.

This leads us to the understanding that a system that is not conscious of its sensory image and its environment cannot be autonomously mobile. Any mobile entity acting otherwise is either not completely autonomous, or is under the direction of a conscious entity. Is it not true that in nature we rarely meet with a non-conscious mobile entity? In nature, the loss of consciousness instantly kills mobility, which could be a pointer to the link between natural consciousness and mobility.

We did say that intelligence systems rise to meet learning demand and therefore the intelligence and consciousness systems of mobile entities are at higher levels and demonstrate higher activity than that of simple static or non-intentional-y mobile organisms. In fact these two artifacts, intelligence and consciousness, constituent of all learning systems, artificial or natural, are intertwined like the snakes on the staff of Caduceus (Hippocrates?)

**Consciousness Driven Learning Systems**

Turing (1) did foresee the major problems in intelligent entity design in his 1950 paper on whether machines could think. He also foresaw the logical conditions necessary for autonomous learning and the approximate path to it, but some interlinking mechanisms are missing and the picture is not very complete. For our design, we reuse Turing's idea of the punishments and rewards mechanism, which he posited as useful in child machine learning. Rather than use Turing's terms, we will use the terms success monitoring or success/failure mechanisms.

We will first develop a simple consciousness based learning system and wonder about its enablers, constraints, and its performance in a natural environment. From our discussion on consciousness we know what consciousness implies; a system image, system sustenance demands and system protection demands (a sensory boundary image and the need to sustain and to protect that image).

Here we create a simple system that uses success monitoring and associated mechanisms; one that is better explained using control system terminology. In deference to readers who may be natural scientists we keep the scope and terms of the discussion as simple as possible.

# Conscious Intelligent Systems

Natural Intelligence and Consciousness – A Learning System Perspective

## Part I: I X I

Our system is a twin loop control system, with an outer loop straddling the inner. We did discuss a simple intelligent system earlier, a simple learning system with an archive and other artifacts, in our basics section. This simple intelligent system will form the inner loop and we will provide a success monitoring mechanism that will form part of the outer loop. This is the learning loop.

This outer loop will be driven by system demands, both of sustenance and protection; so both food demands, environmental demands and environment protection demands form part of this loop. This means that the outer loop contains both the system image for system protection and sustenance demands like food and also mobility control. We can therefore call this the consciousness loop.

This consciousness loop contains a success failure monitoring mechanism that determines if the systems response satisfied the system's internal or external demands (sustenance demands and environmental demands). The success monitoring system has as its reference system protection and system sustenance demands. The demand is passed on to the learning loop, which will respond using either online learning or cognition and response from the archive if prior solutions exist. The success monitoring system will use environmental feedback and tag all responses of the learning loop to a (system or environmental) demand as successful, failed or partially successful (but non-optimal) based on how well the response satisfied the demand.

Partial solutions arise because of response time constraints when this consciousness based learning system inhabits nature like environments. Partial or non-optimal solutions themselves can be marked with a success percentage, so that they straddle an optimality range. We know that natural environments provide variable time demands on the organism, many a time, the response demands are instantaneous, and there is no time available to actually learn/process data. This means that an instantaneous perfect response to the natural environment may not be possible.

After some time, we can see that the learning archive will contain many solutions; some tagged as successful, some as failure and some as partially successful. It is clear that non-optimal responses require further learning and failed responses need repeated learning. In a natural environment, such learning is however possible only when the condition or demand repeats.

Notice however that the system as a whole is however aware of what it knows and what it needs to learn. When a demand repeats, successful solutions, if available in the archive, can directly be delivered to the output. Solutions that require relearning have to go through the learning system for reprocessing.

The availability of partial and failed solution tags in the archive actually allows for earlier and better problem cognition when it repeats, when a partial solution exists in the archive and is tagged for relearning, the system can alert remain and can alert the sensors for a repeat of such a problem. In effect this is a iteration enabled, postponed learning system that can not only learn, but also look for, wait and learn when suitable conditions arise, which makes it a good solution under time constrained response repetitive environments. If the demand recurs and time for reprocessing is available, then these partial solutions can be taken up for reprocessing.

It is to be however noticed that natural environments may at times even deny such chances for reprocessing, perhaps when the environment is as constrained as before or when the relevant archive data is so submerged within the archive that the time available for response is used up in search and cognition. This means that the system has to be always open to the possibility for partial learning. Such a wide prevalence of partial learning demands that reprocessing does not start from first principles every time, learning should continue from where it left off, call it stop and go learning.

Till this point there is little new when seen from a control system angle, this is a simple multi loop control system and higher and more advanced systems see service both on earth, in the air and in space. Patience please!

Notice that in nature, humans and most natural organisms always seem to learn iteratively; improving their responses in kaizen fashion over multiple learning cycles and even multiple lifetimes. Such iterative learning and partial relearning possibilities demands a special kind of learning and archive design as we discuss later. We call it a scaled learning and database design. We will discuss how such a design can possibly aid search, learning, and pattern identification. The system we have discussed now can however do incremental postponed learning, but is not good enough. Some new mechanisms need to be added to improve learning capabilities.

Did you notice that this system contains a factor that is more important to us from the point of consciousness and learning? If you did not, notice that this is the beginning of directed or intent driven learning; the consciousness loop gains the ability to vary and direct the learning demand and the learning process to its current requirements, thus making it a truly autonomous learning system. It can even set its priorities on partially learnt solutions and take up learning on a priority basis. Since these demands arise from an entity being conscious of its own demands and its environment, a self driven intelligent system fitted with a success monitoring loop and system protection demands, and capable of iterative learning becomes an autonomous, conscious, intelligent system.

The effect of directed learning is however greatest on the sensors, orienting them, and supplying them with a focal set of demands that removes their endless dithering and focuses them on what to see. This in turn has a great effect in minimizing the extent of the frame problem. Isn't that interesting? This does however imply some intelligence on the part of the sensor.

We did see that given nature like environments even pattern-based systems are starved for learning and processing time. The severity for mobile complex systems is greater. In such a case, the provision of an additional loop via the success monitoring system can actually worsen the system's responsiveness in the short term. The additional consciousness loop adds an inevitable processing delay in the system, thus enhancing survival risks at least in the short term when the system is under response pressures. We cannot however wish away the consciousness loop; good result oriented learning depends on it, so does mobility, consciousness is but a necessary devil.

We also did see that even for the simplest learning systems, learning improvement with respect to a particular environmental challenge needs recurrence of the challenge. Even during such recurrence, environmental/time constraints could still deny the possibility of complete or improved learning; the entity per se has to wait for multiple occurrences of the challenge before a good quality solution can be achieved. The need for multiple learning cycles is very high.

For entity's that have life spans of hours, days and months, multiple learning cycles certainly look like a luxury. One can escape this conundrum by saying that learning cycles will be spread over multiple lifetimes, something that is true to a certain extent of natural entities. It is clear that even for the most advanced natural entities, such demand recurrence based iterative learning will mean that, with such a system, the prospects of good learning are available but limited.

**Dilemma**

Can we avoid the wait for demand recurrence? Can we improve the learning prospects of such a system? Can we shorten the learning cycle time? Can we improve the learning per learning cycle?

Under response pressures implicit in a nature like environment, given the processing demands and delays implicit in online learning and the requirement for offline processes, it will be advantageous if we can split the entity's processing and response mechanisms. We can let an archive based environment response process run in the foreground and shift learning processes to the background. Transferring the load to an external process or a higher-powered system and gathering back the results will also help. However these are at best possible solutions for artificial autonomous mobile entities alone. Why is this so?

Notice that natural systems are already under learning pressures and that their learning abilities are limited. The effects of morphology, environmental factors, predator prey relationships, and food availability further constrain the quality and quantity of learning available to an entity.

# Conscious Intelligent Systems

Natural Intelligence and Consciousness – A Learning System Perspective

## Part I: I X I

The only positive that the environment provides them, considering all other factors as constant is a little rest time and the presence of low environmental demand periods. In a natural environment, due to various factors like the sun's cycle and the weather, most natural organisms can and do enjoy periods of low environmental demand and some much needed rest. This rest time is generally a time of low dynamic processing loads, learning loads that arise from the environment and from internal system requirements are generally low in such periods.

Learning during such low load periods can be advantageous to the entity. This implies learning load shifting. Is this possible to use this time to learn? Can we shift online learning loads to these rest periods? If learning can be accomplished during these periods, then it will free up the organism to concentrate on online response search and response delivery. We can see that rest period learning, if implemented will result in great advantages to the entity.

However our very definition of consciousness and intelligence precludes this very possibility. The very idea of consciousness and the resulting learning mechanisms arises from that of the sensory boundaries. Any learning demand arises with respect to the sensory boundary and there can be no learning demand that lies outside of it. All data to a consciousness based learning system are sensor-generated data and our learning system is designed and oriented not to learn from (sensor reference free) data chunks but against an incoming sensor based data stream.

Notice that even during rest periods too, the sensory boundaries are not exactly idle, they feed in data, and such data needs to be processed even though response demands may not be urgent. Also processing any data other than online data while being online may result in incongruent system responses to the environment, another risky endeavor. On the other hand a background process looks difficult from the point of view of available processing power and resources.

Let us for the moment recap the demands that precipitate the need for a rest time learning mechanism. The major cause is the systems decisions, forced upon it by the environment, to respond now and learn later. This demand curtails learning time, can interrupt search and creates sub-optimal learning solutions and system responses. Given a certain processing power, the ratio of sub optimal solutions to complete solutions is going to rise with increase in system complexity or environment complexity.

Notice that right from the simple intelligent system to the pattern based system the demands for rest time learning do grow exponentially. We saw in an earlier section that pattern identification and archival management demands may necessitate that the entity finds times and places of rest, safe from extreme environmental demands and challenges.

The demand for iterative learning and failure relearning that rises from the consciousness based learning mechanisms we have just discussed also increase the felt need for a time of rest when all this processing can be carried out. Without the presence of rest time and associated rest time processing mechanisms, an archive-based online cognition and response/ offline learning strategy will fail.

If rest time is available then processes like archive management, deep search and pattern identification processes can be executed without much system modifications since their process demands are generally limited to the availability of uninterrupted processing time and resources. However as discussed earlier, rest time learning remains a problem.

As we noticed what we have now is a learning mechanism designed to learn against a sensory boundary based data stream. Even during rest, the existing sensory boundaries are busy. The old data needs to be replayed for the rest time learning mechanism to learn from it. In the absence of a sensory boundary we will need to create a new learning mechanism to learn from that data. We may not have the resources to create a parallel learning system.

Can we use the available facilities to resolve our problem or should we create a separate learning environment and affiliated learning processes?

# Conscious Intelligent Systems

Natural Intelligence and Consciousness – A Learning System Perspective

Part I: I X I

**Solution**

Remember that we have a sensory data archive and that data from the sensors go into it in movie frame fashion. Reversing the movie frame can give us data recall. Remember that sensory interfaces bring in these data. If these sensory interfaces could be mimicked in memory, stub like, we can then recreate the entity environment interface. We can now replay the archived movie frame like data sets through this (offline) interface to let the learning mechanisms learn from it. As far as the learning mechanisms are concerned, it sees an actual entity in an actual environment and can therefore learn from it. However there is a small hitch.

Notice that the offline learning mechanisms are generally concerned with reprocessing or relearning loads. In such loads the solutions that the entity provided at the time of actual occurrence are also available. In general such a review based learning process will be slightly different from the normal online process. How will a learning system even if given time, learn from such solution embedded data? We will take up this problem in a later section.

Notice that very presence of such a rest time learning process means that with a little resource sharing the entity can use its lightly loaded rest times to learn/relearn, while responding to the lighter demands of the environment. When interrupt levels from the environment rise and there is a demand for quick responses, then the rest time learning process can be aborted and the entity can go back to proper online activity.

Notice that in theory such rest time learning can be really fast. The happenings of the day can run thro the entity's memory mechanism in less than half the time. Why should this be so? You do know the answer, however to be more explicit, let me say that the rest time learning process runs at processor speed and not environment speed, not being dependent on the environment for reactions saves a lot of time.

It is clear that such an offline learning mechanism, if available, would be a great boon to the entity. It can review its actions and environmental reactions and learn from it at its own pace at a time when it is not besaddled by environment response concerns. Such a mechanism increases learning time, improves learning, and can increase the learning prospects of an entity in a natural environment.

Notice that in the usual case (except for the predator) there is practically no response time pressure on the rest time learning mechanism. The single biggest inhibitor of good learning quality, the time constrained learning demand, can be countered by such a rest time learning process. Rather than relearn or review the entire happenings of the day, we can reduce rest time processing loads by choosing to process only directed learning demands. The success monitoring mechanism and the consciousness loop can create a priority ranked list of learning demands and direct the offline consciousness based learning mechanism to recall relevant events and data and learn from the data. Iterative learning and Failure relearning become easy and the system will have a trial response ready in its learning archive, next time the demand recurs. Whenever the entity is free, then such free time can in theory be offline learning time.

The most important factor is that while this arrangement improves learning, it does not demand extra processing power or radically new architectures. The same learning mechanism can be reused for such review and the results of offline learning can be added to the same learning archive, seen that way this is an economic design that reflects Thatcherian prudence. From an engineering point of view creating a mock interface is not an easy job, but far easier than creating a separate learning system. If such a load shifted, rest time learning mechanism sounds like a fulfillment of a time constrained learning system's wish list, wait till we encounter the problems.

Notice that in recreating the sensory interfaces, we are recreating the entity in memory, albeit in a limited sense. If you remember our definition of simple consciousness as awareness that rises as a result of the sensory boundaries, then this recreation of the sensory boundary essentially duplicates simple consciousness. A shadow like simulated entity comes to dwell in memory. The data replay or learning review process runs using this simulated entity as the primary actor to interact with the simulated replayed environment. Therefore in doing such data replay we not only have environment data recall and environment interaction history recall but also entity recall.

During offline learning, the entity has to be aware of this simulated entity, its learning processes, and its results. Such awareness is best illustrated by the example of a stage and an audience. It is as if the basic conscious entity is in the audience, watching its alter ego, the simulated entity, performing with other actors on a stage. (This in effect puts the actor in the foreground and the audience in the background) Such awareness is essential to avoid environment incongruent responses that may arise out of response tangling.

Response tangling arises because the learning mechanisms are reused for offline learning, and its results go into the same learning archive. It is always possible that the result of an offline process is wrongly routed to the system outputs and confusion and environment incongruence can result. The entity will seem to exhibit mad behavior. Such response tangling can be extremely risky and even fatal to the entity. It is therefore absolutely necessary that the entity be aware of these offline learning results as separate and arising from its rest time learning mechanisms. It is also necessary that these processes and results be corralled away from the online response mechanisms.

However with such an offline data replay based learning process, it is clear that the entity has a higher learning potential. The quality and quantity of such learning will be dependent on the amount and quality of rest and safety the entity can enjoy. With a rise in learning potential, the entity can even seek higher processing load by colonization or migration.

Notice that the very presence of a shadow entity and offline learning mechanisms offers some breathing space to the harassed entity and its online learning mechanisms. Learning pressures come down; rest time processing also ensures a better and more complete quality of learning. This leaves the basic consciousness based learning mechanisms to concentrate on faster archive based search and respond processes, confident that the offline learning mechanisms can handle any new learning loads. The presence of such benefits from offline learning would mean that the entity would be motivated to find the rest time and the safety needed for such processes so as to maximize its learning potential and enjoy the chances of a more relaxed online life.

Rest time is rarely a great problem for natural entities; the diurnal nature of the day is a great help, other rest times do appear in the daily life of any natural entity. The problem then boils down to the level of entity safety, higher the safety during rest time, the better the entity can learn. The combination of good rest and safety for natural organisms depends on a lot of implicit and explicit factors like morphology, lifetimes, predator prey relationships, food availability, evolutionary paths etc, in short the evolutionary and ecological history of the entity and its species. Our present knowledge about processing power availability and processing architectures in natural entities is too low to make an educated guess about these factors and their level of influence.

Therefore, for purposes of argument, let us ignore individual species capacities and foresee the scenarios that arise from a combination of these three (offline) learning critical factors; rest, safety, and processing power. If we suppose that these factors have binary states, eight combinations become possible; some of which lead to extinction, some to stable online systems and some to systems that need offline architectures. A combination of 000 is a call for extinction, the combinations 100 and 010 are exceptionally risky and may not survive, the combinations 001,011, 101 and 111 can lead to pure online learning systems. That leaves the combination 110 as the one that may need offline learning processes.

Pure online processing power may look like a luxury in nature, where the processes of natural selection imply that a demand for extra processing power may take generations to satisfy. It is well known that processing mechanisms tend to consume a large part of the entity's energy. It can therefore be presumed that the processing mechanisms will be constrained to stay within the energy minimum that the system can support, since these systems need to be functional even at such minimum energy levels. Then there are other systemic and environment based negatives that act to ensure that processing power availability stays within processing power demand. Despite such negatives, it is still possible that some natural entities end up with an excess of processing power, even if temporary, mainly as a result of evolutionary trajectories.

For purposes of argument, consider a condition in the evolutionary history of the organism where there was a sustained demand for a rise in processing power. Such demands may have risen due to factors internal and external to the entity. An improved sensor may demand more processing resources, so could variant environment conditions. Let us say that over generations of natural selection, the entity and its species rose to meet the increased processing power demand. Let us now allow a long period and multiple generations to pass by after which we say that the original conditions that demanded such power increases have died down or disappeared. In such a case the entity will be left with an excess of processing power. If we use the logic of natural selection, we can safely presume that there will never be complete synchrony between processing power availability and the rise and fall in processing power demand. The response of the entity and its species to any increased or decreased demand generally takes multiple generations to satisfy. There can be an interim condition where the entity and its species enjoy an excess of processing power.

In such a case, like a man newly awash in money, the extra processing power will be directed towards new baroque or functional targets. The entity and its species may act to colonize new environments and thus take on new learning loads, alternately it may undertake to fine tune its courtship strategies or optimize other systemic responses. If this extra processing power contributes in any way to the success of the entity or its species, then despite the fall in actual processing power demands, we can expect natural selection to retain this extra processing power.

A stronger case for the presence of extra processing power comes from entity or species dwarfing. Environmental and systemic factors may force entities or species to take smaller body shapes to keep in tune with environmental or other demands. It is logical and also well known that brain sizes are proportional to body sizes. If we reuse our earlier logic, we can see that such downsizing can also lead to a condition where the entity and its species enjoy a temporary spurt in processing power, and so on…

The author wonders if (other than for genetic reasons) such size reduction was a primary trigger for those natural organisms that have come to enjoy better brain to body size ratios. He is not aware if there is any prior evidence to support such a premise. Such an engineering based logic may not hold true in natural realms. However if there is evidence to support such a premise or if we can assume that such a premise is true, we will show in a later section how it is possible to use such a premise to explain hominid evolution.

While the above case for excess processing power needs such an arm-twisted explanation, a case for lower processing power can be much more straightforward. The addition of processing power is costly and resource intensive, so one can expect that processing power supply will generally stay below or just equal to processing power demands. However a condition of too little processing power could also be risky to the entity and the species. Given that delays are inherent to processing, too little processing power can make these entities sluggish and unresponsive and thus expose them to environmental and systemic risks that may lead them to extinction.

However entities or species with processing power shortages can still maximize their learning potential if they have the rest and safety to afford offline-learning mechanisms. This is the ' 110 ' condition we discussed earlier. Rest and safety are critical to offline learning processes since offline learning mechanisms have to share processing resources with the online mechanisms. A lack of safety and rest would mean that the entity would not have the time and flexibility to allot processing resources to this offline mechanism. Even when used, these offline mechanisms are susceptible to quick rollback calls, which in effect debilitate learning possibilities.

Since rest time is generic and available to almost all entities in nature, it is the level of safety that determines the level of offline processing possible for the entity. In cases of good safety, like say a good availability of shelter and rest, the offline processes can increasingly take over online resources. In cases of exceptional safety and where high demand for offline processing resources exist, it would even be possible to shutdown the online consciousness mechanisms or reduce its processing power requirements to a minimum. This allows the offline mechanisms to garner almost all the available processing power for offline processing. Does such a condition correspond to sleep?

Many natural organisms sleep; naturally many scientists have wondered if sleep time corresponds to learning time and have sought to study them from such a perspective. Unfortunately the real evidence has not yet tallied up to strongly support such suppositions. The jury understandably is still out on the evidence. To us viewing the problem from an engineering point of view, sleep looks like the ideal choice for an offline learning process. It is not necessary however that this be true from a biological standpoint.

Other reasons for sleep are presently being conjectured. For instance energy conservation looks like an obvious and logical choice, however notice that even during hibernation, there is evidence that sleep periods are separate and demarcated. The presence of the circadian and similar environment based rhythms is also evidence that most natural organisms have a great level of (inbuilt) awareness of their larger environment. It is possible that through the course of evolution low load higher safety periods were identified and earmarked for offline learning. The best we can do is to wait for more evidence from science to support or reject a conjecture that sleep times are also learning times or that sleep arose primarily for learning purposes.

However if the sleep conjecture is correct, it can be seen that more sleep naturally translates to higher learning potential. Higher the processing load and higher the processing resource constraints, higher would be the requirement for sleep. Better the quality of sleep, better the learning would be, within the given processing power capacities. We can see that a facility for uninterrupted sleep can multiply by many folds the processing loads that a system can theoretically process.

What happens to entities with little safety or rest? One presumes that such entities will need to do some sneak-in offline processing, running a mix of offline and online processes when rest and safety allow them to. Such an admixed process, though risky, can be a boon to such entities that are starved for processing power. During such periods, online response may be a little compromised; a decision to run such offline processes online demands a cost benefit analysis. In the natural kingdom, the rest and safety available to an entity and its species are dependent on the entity's position in the natural hierarchy.

Prey would find such an admixed process difficult to implement, because the need for high alert means that they would find it difficult to relax and allot time and resources to the offline mechanisms. The threat of rollback is perennial and high. However in the case of most predators, such an option can not only be implemented but can also be useful in a manner we are yet to consider. The presence of a offline learning mechanism combined with the (high priority) need for hunting success will tend to turn the offline learning mechanisms into an instant replay and review facility. Such an offline review facility will allow quick reliving of the event history and learning from it. This will allow the predator to fine tune its strategies on the run and aid their run time intelligence.

We see that rest and safety remain the most critical factors in the implementation of an offline processing mechanism. If rest and safety are available, then the target entity's learning potential can be maximized. How can such a theoretical maximum be achieved?

Consider a case when the entity has the facility to run an undisturbed offline process during sleep and an admixed offline-online process during wakefulness. With such a strategy the learning mechanism enjoys the maximum possible processing power and the maximum offline learning time. This is the condition where its learning system achieves its highest possible learning potential.

It is natural to ask whether all this learning load involution and sneak in processing really necessary. Prima facie, no natural entity seems to be so harassed and point bent on learning or data processing. What are the learning loads that necessitate such learning process involution sneak in processing and sleep? Where are the processing power constraints that we talk of?

# Conscious Intelligent Systems

Natural Intelligence and Consciousness – A Learning System Perspective

## Part I: I X I

The most straightforward answer would be we do not know, nor does current science. We are just beginning to get a hang of the learning processes of animals and man. However from an intelligence system design perspective we can say that such loads exist and that an offline solution that corresponds to rest, safety, and sleep looks like a workable solution. Notice that offline processes are not confined to offline learning, it also includes processes like archive management and pattern related processes that range from identification, cognition, pattern solution creation/play out and so on.

Notice that our discussion until now has assumed that the inner learning loop is inhabited with a simple learning system. When we replace this system with a multi sensor pattern based system, we can see that online processing loads are not only high, which in itself can increase offline-learning demand, but also that archive management jobs have multiplied, and pattern identification becomes distinctly more difficult. Mobility adds to these increased processing demands, both online and offline. Notice that we ignore a lot of data during mobility. Such ignored data also wait for offline processing in the archive. Mobility also implies sub intent devolution and sub intent reassembly, this also increases processing loads. The moving target of environment comfort also brings in its own learning loads as the entity looks for measures to do more with less, for instance reduce food availability variability, reduce predator attraction, improve prey location and so on. Then there are the small and large environmental variations that are part of any environment; all natural entities need to keep pace with such learning demands that arise from environment variation. All natural entities also seem to take risks and learning to the limit of their processing loads, within their generation and life spans and this is what keeps the species alive and moving. Even otherwise learning resources are used up in the never-ending one-upmanship with members of one's own species in the race for the survival of the fittest.

The author avers that the present trend towards neuroscience based brain studies will increasingly bring evidence of such learning loads, resource constraints, and offline processes, in many organisms including man.

**Mind**

In our discussion of the offline or rest time learning mechanism, we said that the entity should be aware of the simulated entity, the associated secondary learning process running on the simulated entity environment interface and its results. Such awareness of this secondary learning environment will allow it to differentiate itself from the entity and corral its results and processes so that they do not clash with the online mechanisms.

If we can map our offline learning mechanisms (with its entity recall, environment recall and environment-entity-interaction-history recall attributes) to natural learning systems and organisms, say humans and animals or say fruit flies, we can figure out their equivalents. OK you already guessed it!

Let us call the simulated learning theater as the mind, the arena in which the simulated entity interacts with a simulated environment and learns from it. Let us call the learning processes that run in such an environment as the mentation processes and let us term thoughts as results of such mentation processes. We know that when two processes run in parallel in a system, then the processes need clear system identifiers; let us identify the shadow entity as "I". The sense of I helps the main entity identify its shadow entity as distinct and separate from itself. Our basic consciousness mechanism is aware of this I as separate but as a coarse reflection of itself.

This equivalence between the artifacts of the offline learning mechanisms and the natural intelligence system is a basic assumption (presumption?) of this discussion. We assume that our sense of "I" denotes a proxy entity that inhabits a proxy-learning environment, which is our mind, that offline-learning processes correspond to our mentation processes, that thoughts are results of mentation processes and that the basic consciousness is aware of the proxy entity as I.

From our perspective, we see the mind or simulated consciousness as an offline-learning environment necessitated by high offline learning loads combined with the availability of rest periods and comparative entity safety. More the learning load, more the requirement for the presence of the mind mechanism! More the safety and more the rest, more the time such an offline learning mechanism can occupy the brain of the entity. More the time the mind occupies the brain, better the learning capacity of the resource starved entity.

If this equivalence is correct then we can see that the presence of the offline processing and learning architecture can allow natural organisms to take learning and processing loads that are simply not possible without such architecture, given processing power and other resource constraints. If the presence of consciousness marks the beginning of directed or intentional learning, the presence of an offline learning system running in simulated consciousness optimizes the learning process. The entity's basic consciousness and its mirage like sister, the mind share the learning load and reduce the resultant learning pressure on the system. Such a bi-conscious system can handle higher learning targets and pressures if it has sufficient rest time and environment safety.

During wakefulness and periods of high alert, the mind is constrained to give place to the online basic consciousness mechanisms for environmental response purposes. This online response requirement cannot be wished away because man or any other animal that has a mind is basically a mobile entity. Basic consciousness, we did discuss, is a pre-requisite for mobility. During sleep, when our environment safety allows us considerable leeway when compared to the animals, when mobility demands are absent, our mind can shift to top gear.

**Presence of Mind**

Do animals have minds? Do insects have minds? Does man have a mind? If so, why does he need one or how did he come to one? To answer such questions, one needs to learn and understand more about the learning loads of a natural entity. Such data being not available, we perforce need to speculate. Our present understanding of the process however allows us to offer some logic to our speculations. In the absence of real scientific data any explanatory process is inherently speculatory (and must be taken with the usual bucket of salt).

If mind equates to offline processing then one can see that for entities that have the facility for sleep the offline processes would run during sleep periods and will be finished when they wake up. During such a sleep process, the cognizance of binary consciousness is not necessary and the entity will not be aware of this learning process running within it, except when it dreams. (Perhaps as a result or part of a timed/regular keep awake interrupt) In the general case, the entity only knows that it needs to sleep and that after sleep problems that looked highly intractable prior to sleep look reasonably soluble.

Mobile entities that have migratory and colonization activity will need offline processing time and offline processes to update their databases. Even here if such updates can take place within sleep or deep rest, cognizance of the process is not really necessary. The presence of mind can therefore be ephemeral in natural entities that are gifted with sleep. If and when they sleep, they can be said to have offline minds that work as they sleep. The level of sleep is then an indicator of their presence of (offline) mind. If offline minds come on only during sleep time a cognizance of I may not be even necessary. The sense of I would make brief appearances as they go into sleep, when they dream and when they exit out of sleep.

We did see earlier that some entities like predators would benefit from having cognizable minds, even online. Let us say that they may need online minds, even if ephemeral. On the other hand, for prey; the attraction of improved learning is at best dicey. In short, natural conditions may not allow most natural entities the luxury of online minds, while offline sleep based minds look a distinct possibility. Such a broad based generalization may not apply equally to all entities, however our present lack of knowledge about these entities learning loads and processing power leaves us with little option.

Man is not exempt from the natural kingdom. Notice that emergency conditions and exceptional alert conditions do debilitate his sense of I. When emergencies arise, humans are also forced to fall back on their instinctual mechanisms. In such situations his sense of I goes missing, the pattern of individualistic activity he normally demonstrates vanishes; he becomes nothing more than an instinctual animal. In such situations he can neither see himself nor understand the logic of his actions when such actions come under subsequent review. The best he can say is that he was not acting self-consciously.

The last paragraph implicitly assumes that man has and needs I, that he has online mind. Why does he need one? What triggered the rise of the human mind? What makes him special or his needs special from those of the other animals?

The processing power requirements of any entity or its species are in broad terms dependent on their evolutionary history. At some point in the evolutionary hierarchy there arose a remarkable intersection of processing power, rest, and safety that heralded the arrival of man. The author feels that this remarkable conjunction of processing power, rest, and comparative safety can be seen even today with the advanced primates.

If you look at the advanced primates, they are largely herbivorous, largely non-hunters and on the whole, are not preferred prey for any predator, despite their having the morphological and intelligence faculties a predator may need. Whatever the reason for this predator prey food chain drop out, this seems to have provided them with considerable environment safety and ample rest time. They also seem to have a predilection for comfortable sleep. Even Darwin (2) in his comparison of the mental powers of animals and man was tempted to make an observation in this regard.

"The orang in the Eastern islands, and the chimpanzee in Africa, build platforms on which they sleep; and, as both species follow the same habit, it might be argued that this was due to instinct, **but we cannot feel sure that it is not the result of both animals having similar wants, and possessing similar powers of reasoning.**" - Chapter 3 Descent of Man (**emphasis author's**)

Darwin's remark allows us to posit a view that may sound radical. We did say earlier that sleep arises when ample safety and rest conjunct a lack of processing power. This makes us wonder if the advanced primates are actually starved for processing power. As of now science is still not clear as to what triggered the hominid speciation process from an ancestor common to both the advanced primates and modern humans.

However notice that in the evolutionary branch off, based on the fossil evidence that we are presently privy to, the hominids actually started off with smaller sizes and bipedalism. Did smaller sizes place these initial hominids on an equal footing with other prey? A freak thought suggests itself; did bipedalism itself arise as a response to being chased as prey? What caused the downsizing? Did this downsizing lead to better mobility? Did such increased mobility lead to better vision?

More important to us is the question as to whether this downsizing gave them a processing power advantage, in line with our earlier suggestion of better brain to body size ratios? Did that small extra make all the difference? Did the improved brain size to body size ratio trigger the start of the hominid rise? Did he find the extra processing power useful to resolve (may be even belatedly) small problems that his genetically close brothers, the apes could not solve? As a species we are not natural carnivores, that much is certain, also we have inbound fear of many of the other predators which is in itself unusual for a predator.

We know that such speculation hangs on a limb too thin, this is a subject on which no clear answers presently exist or can be given, however there is clearer evidence that the hominid brain added more than a kilogram of weight to the head during its one million years of evolution from prehuman to man. Science posits an evolutionary hierarchy that led from the apes to man, the lines are however not still clear and form the subject of considerable debate.

Let us look at it from our learning system perspective. We see that processing power increases can be triggered by both internal systemic needs and external environmental needs. When we track hominid evolution, we see a picture of increasing brain size. If we assume that most of nature uses processing equipment that are close to uniform, then we see that processing speeds would remain practically the same, there seems to be little indication that man's nervous systems or its processing are quicker than that of the other animals. So when brain sizes and learning loads increase, it is obvious that given the same processing speeds, with a more complex learning architecture and increased processing loads, more learning time is necessary. While we do not know the actual learning demands, a comparative look at the advanced primates sleep requirements tells us that the offline processing slots are practically full; they all need their eight-hour sleep quota. This means that any additional learning time demands necessarily have to cut into online process time, unless the environment permits higher sleep durations or the brain mechanisms speed up the learning processes.

When one posits such an online cut-in of the mind into hominid evolution, (based on what is known about hominid evolution now), one sees the possibility that the early hominids would have found their online rest times being increasingly taken over by their mind mechanisms. With each increase in brain size, the mind needed to increasingly cut into online times. The final point of such a cut in process would be the almost parallel presence of the offline and online processes, in other words the mind and the basic consciousness mechanisms shuttle or oscillate between themselves incessantly, each giving space to the other. Notice that this matches the theoretical conditions that we discussed earlier on, a condition where the learning potential of an offline-learning enabled entity is maximized.

The author posits that this is the condition where the modern human species finds itself in with its incessant shuttle of the mind and its basic consciousness. Man seems to have come to such a maximized state of learning where an admixture of online and offline learning processes consumes his day, while sleep consumes his nights. He has both online and offline minds. Notice that there seems to be little space for mental Superman; the learning slots seem pretty occupied! Notice that this puts a theoretical full stop to intelligent system evolution on this evolutionary branch, which is perhaps why evolution did not continue to make Superman.

It could even be speculated that this moving in of the mind to a more active online presence is what triggered the dietary move of the hominids from major herbivore to major carnivore. This is because they now had a review mechanism similar to that of a predator; they could outthink their prey and fast. For the earlier and later hominids, the increasing presence of an online mind perhaps helped change the equations of the traditional food chain.

Being a predator changed his status in the animal kingdom from sometime prey to someone who needed to be feared. If his size did not deter them, his group sizes would, if that did not deter them, there were the tools that were getting rather fearsome. The move to a carnivorous diet would have also meant better food supplies and can probably account for their subsequent increase in sizes. It was perhaps this dietary change that gave them the impetus and safety of migration and colonization; they were no longer dependent on a vegetarian diet. Another concurrent process happened here, one we cannot explain how, his increase in fertility, the links are still not clear. However the author speculates that there is more than a casual link between learning systems and fertility, given our earlier supposition that reproduction is essentially a pattern passing process. He wonders if learning stresses debilitate fertility.

It can be speculated that the forward movement of the mind into online time and the resultant species success brought a forward loop into operation, a loop that ended with modern humans, who now have the requisite safety and rest to have their mind flit in and out of their basic consciousness within an eye blink, literally. Before they got comfortable with it, possibly over a couple of million years, such a creep in would have been very unsettling to the earlier hominids, particularly for the earlier ones, having to contend with a shadow entity that arose in them at rest times and proceeded to provide solutions to problems that they had just faced. Once it grew enough for them to be comfortable they must have craved this shadow entity and even advantageous to have him around. Perhaps the very human attraction for intoxicants began with the drive to let the basic consciousness relax and let the mind speak.

Such speculation may not be true, but definitely interesting. It could be argued that our basic propositions are wrong but notice that even in modern humans, there seems to be a correlation between body size and brain sizes, lesser the average body sizes, higher the average brain sizes, if not intelligence. There also seems to be a link between intelligence and carnivorous behavior across the animal kingdom. There also seems to be evidence that links the increasingly human reliance on vegetarianism due to his agricultural successes to a net decrease in average human brain sizes over the last 10,000 years. Smaller sizes across the animal kingdom have implications for group size also and there is increasing evidence that there are parallels between brain sizes and social behavior. Whatever be the major impetus, it can be presumed that it is the enlargement of the online mind that necessitated the further enlargement of the brain and carried evolution further on till man was far ahead of the advanced primates. (Some of the above conclusions remain under debate in the science community; we do not wish to take a position on them. Much of these quoted data/conclusions arise from general science articles in various media and do not refer to specific science papers)

It is natural to ask as to why other and earlier predators did not rise up to man's state. For explanation we will have to fall back on our evolutionary history, our learning demands and more importantly morphology; bipedalism was the first and most important enabler and differentiator for the hominid line. We presume that the speciation process that diverged man from the apes also provided him with a little extra processing power that reinforced and grew to human scales over evolution. We presume that the mechanism of the mind must have risen early on in evolution, but has occupied the echelon in man. The author suggests that during wakefulness the human brain shuttles between mindlessness and mindfulness in the blink of an eye. He wonders if the eye blink is actually a state change signal of the human brain as it shuttles between online and offline processes.

I

The presence of the (online) mind creates a binary consciousness in any entity it inhabits. If the concept of self-consciousness (def: aware of yourself as an individual or of your own being and actions and thoughts) indicates that man is aware of himself as an entity independent of his body, from our perspective we can explain how such a feeling can arise.

Offline processes need not be unique to man and so man is not necessarily the only mindful entity, even an animal can demonstrate mindfulness and binary consciousness if not self-consciousness. Even online minds need not be unique to man; the best place to look for them might be in the larger predators.

Man's awareness of this self-consciousness has challenged him over millennia; the concept of soul and all philosophy, even religion seems to have risen from this seemingly separate entity that inhabited the body. An entity that baffles neuroscience by refusing to show up in those brain scans! The stranger fact is that humans have been aware of the mirage like nature of our self-consciousness for millennia now; the proponents of quietism were perhaps to notice this. The quietists have over millennia converged on one single question. Who am I?

In answer many of them have persistently claimed that our sense of I is a mirage and that man's actual consciousness resided below this mirage. Across time, sects, and religions, this seems to be their only common claim. In consequence, many of them consistently refused to identify themselves by names and used the third person to refer to themselves, rather than say my eyes; they used the format, this body's eyes. This is not to claim that the quietists are right/wrong, but to point out the historicity of the question and their answer.

"Whom are you seeking?" asked Abu Yazid the Sufi. "Abu Yazid," replied the man. "Poor wretch!" said Abu Yazid. "I have been seeking Abu Yazid for thirty years, and cannot find any trace or token of him."

Other philosophers, scientists, and common people have also tried to make sense of self-consciousness, however the Descartian proposition of "I think therefore I am" is in many ways a scientist's attempt to resolve a pure senses based mechanistic argument, that man is a sum total of his senses. In the conventional interpretation, if the author's understanding is correct, Descartes' assertion is implied as a dualistic argument, that man exists as an entity independent of his body.

The idea also implies that humans are different from animals because he is able to think of himself as an entity different from the body, an entity divergent and existing independent from/of the senses, from such a view animals are automata, unfortunately the mechanistic view intended originally for man gets hoisted on the animals. Apologies if the author understanding of the Descartian proposition is wrong!

From our newly arrived perspective, we can now propose that the Descartian "I" is neither the real "I" nor the entire "I". Notice that we are not actually born self-conscious, as children we acquire self-consciousness by correlation, by correlating between our thoughts and our bodies, just like baby chimps and dolphins also seem to be able to do. We are actually born binary conscious.

Over time and evolution the permanence of our learning loads and the need for the almost permanent presence of mind have morphed this binary conscious mode into a linked consciousness mode that allows correlation between our thoughts and our bodies. Thus do we arrive at self-consciousness! We should perhaps prefer to be called self-sentient to retain our observation that basic consciousness, binary consciousness, and self-consciousness are all dependent on sensory data.

One of the reasons for the rise of self-consciousness would have been to reduce the schizoid discomfort that binary consciousness can induce in the entity, particularly en route to its maximization. In the common lingo, with a binary conscious offline mechanism, one would expect to have voices speaking in our head as the brain shuttles between the results of the online and offline process. Self-consciousness arises as an attempt to reduce the discomfort that the presence of a shadow man can create by correlation between the real man and his shadow. Another possible reason for self-consciousness to arise must have been to rein in a runaway brain. A learning system endowed with good processing power, good inferential processes, rest and safety can extrapolate events and sequences to an extent that they get out of reality and some reining in must have been necessary to keep its solutions relevant to the task at hand. Self-consciousness perhaps arose as a solution to avoid such a runaway. If these are true, one can expect natural selection to have acted in favor of self-consciousness over plain schizophrenia.

We can presume that over evolution, this self-identification and correlation process has acted to create and strengthen our sense of "I". In reality the "I" denotes a proxy entity that resides in the brain and gives identity to the mind, it is actually a design artifact that allows for the presence of an offline learning mechanism.

We did see earlier how the presence and activity of the offline mechanisms tend to push it into the processing foreground, while the basic consciousness based online mechanisms watch in the background. We also did see how the active learning activity of the basic consciousness mechanisms are pushed on to the offline learning mechanisms to allow the basic mechanisms to concentrate on archive based cognition and response.

These are likely scenarios in the evolution of the human mind and self-consciousness. In humans, due to his inherently high learning loads, the mind covers his existence much like a blanket. Even while being a proxy for the main entity, it consumes most of his living time, both during sleep and wakefulness. We are actually more I than not. So what if it masquerades as the real thing? It has every reason to.

**Movies in My Mind**

We know from our earlier discussions that the presence of a success monitoring system is the key to good learning. It is obvious then that the offline learning system, if it is to exhibit good learning, also has to be equipped with one. The real environment forms the feedback mechanism for the online learning process, what could be its counterpart for the offline processes? One can see that the design of an offline learning system is not simple and straightforward.

Learning from a data review process has another problem, which we identified earlier on, the presence of the entity's earlier response. Even though the job of an offline learning system is to improve existing solutions, how will a system learn from such a result-embedded data stream? We certainly need to better understand the entity's solution generation process. Rather than start from first principles and sidetrack this discussion, we will assume that the pattern based learning system generates a response to the environment and our job is to better this response by reviewing the data and the solution.

In our initial discussions, we did refer to proactive behavior as the third solution that a natural autonomous entity uses to reduce dynamic learning loads. We also said that proactive behavior arises out of a marriage of consciousness based intent and the pattern-based system. We said that the time gap between environment cognizance and entity response implicit to pattern cognition systems allowed the entity to demonstrate proactive behavior.

Proactive behavior means that entities tend to utilize the time gap to probe the environment with trial responses to see which gives them the most benefits or helps attenuate risks. Much of the natural behavior we see is actually proactive behavior; a pure online response behavior arises but rarely and is extremely stressful to the organism.

All natural organisms probe the environment for response beneficiation. Such probing cannot be rash and risky; it should be tentative and should contain a provision for retreat should the solution fail. Like a kitten's tentative paw at a ball of twine when it first encounters it, the entity needs to paw at the environment, always ready to reverse its actions if they are risky or when they do not meet system requirements. This can best be done when the probing system has at least an idea of the response it can expect from the environment in reaction to its probing action. How can such knowledge and such knowledge-based solutions arise?

An online system generally generates a solution based on its reading of the problem. The quality of the solution however depends on the time constraint (and environmental risks/demands). With an offline system, such a time constraint generally disappears or is relaxed. This does allow for a better quality of learning. A pattern identification process also relies on the presence of an offline learning system to generate a solution to the identified pattern. Here the presence of the entity's online response embedded in the data stream can help the system compare its new solution with the entity's old solution and replace it with a trial solution that can be tried next time the challenge repeats. This trial solution is the base of environment probing and all proactive behavior.

The presence of the offline processes means that solutions to environmental challenges do not emerge hot from online processing; they are generally derived from the inherited archive or are the result of offline processing. If we can track back to the base system that starts with a pattern identification process one can see that initially the systems solutions are mini solutions, solutions that provide specific answers to specific environmental challenges. As the pattern identification process gathers steam these mini solutions are stitched together to form a response to a pattern. Even the most stable environment undergoes micro changes, which means that the entity has to do a lot of collation, stitching and separation of the micro responses to keep in tune with environment demands. In fact most of a natural entity's lifetime is spent in such micro response collation on demand. Good solution chains tend to get reinforced over time and repetitions and are written back into the pattern-based database. The initial combinations are necessarily carpetbagger like; design and aesthetics arrive over time.

The entity knows the reactions of the environment to its micro solutions. This allows it to guess the environments collated reaction to a solution chain. This means that before the solution chain is delivered to the output the entity has an approximate idea of the way the environment would respond to its latest action. Over time and generations, most environmental challenges and their responses tend to take place within such prebuilt knowledge frames. The result of entity activity and environment response either reinforces or questions this knowledge frame. The everyday variance of the environment is reflected in the flexibility of this knowledge frame, it tries to be inclusive and only when too much divergence occurs does the frame come under review.

The presence of an offline mechanism and the knowledge frame allows the entity to safely probe its environment. If the improved trial solution fails, the older solution can always be reintroduced. In the natural environment this kind of probing is common. Solution improvements could be hard to come by, but all this probing is not wasted, once a solution is learnt, we are one up on the environment and have just increased our comfort level. A rash probing action is necessary only when the environment response does not comply with such internally generated solution frame works.

Notice that the presence of the offline learning system and the archive alters the very approach of the entity to environmental challenges. Notice that with such processes the entity is tuned to catching environment standouts. It is tuned to catch features and challenges that do not conform to the norm. Most of its usual responses are beneficiated responses; the act automatically returns the maximum possible benefit in the given circumstances.

From such a perspective we see that much of natural behavior is proactive behavior or the results of a sum of earlier proactive behavior. Much of the complexity of natural behavior and activity can be explained by the concept of proactive behavior. Such behavior cannot rise out of systems that aim to be purely responsive to the environment; it can rise only out of intentional systems that aim to gain maximum efficiency in their interactions with the environment. Most natural entities are then like Jeeves, always thinking of the many ways they could persuade their master, the environment to follow their wishes rather than the other way round. Man is as good an example as any.

The rat, which carves out a home in the ground, or a squirrel that stores nuts may be doing so out of his pattern based memory, but the rat, which first figured out the solution, had to do some real work of probing and knowledge frame correlation. Later accretion and improvements arise out of iterative learning and the final product may be far removed from what the original intent was. Birds nests, courtship dances, hibernation, many activities which evoke in us wonder all probably evolved from simple practices, much like the tendency of the chimp and the orang to arrange their resting places in preparation for a good nights sleep probably led to down pillows and spring beds and an inability to sleep without air-conditioning.

Our initial problem that sent us scurrying away on trying to understand the solution generation process was the problem of trying to build a success monitoring system offline. Building solutions for a real environment and getting them tested is nice and simple, but what will happen if it goes really wrong? It is possible that the proposed solution carries more than its share of risk, would it be worthwhile trying it? Many possible solutions can arise for a problem; will we get the time, space, and conditions necessary to try all of them? Surely the need for a real time environment and concurrent requirements are a dampener to the system. What can the system do?

It could be beneficial if one could create an offline-based success monitoring system rather than always rely on the environment to provide an online success monitoring system to test the results of offline learning. We already know that we can recreate the environment entity interface and its interactions. We also have a large history bank that contains previous trials and the environments responses to it. Can we use this accumulated knowledge to build a simulation platform to test the results of offline learning before such solutions can be offloaded to a real environment?

While we reserve a detailed discussion on the design aspects of such an offline success monitoring mechanism and its testing environment for a further paper, the simpler way to understand the offline success monitoring idea and its consequences would be to consider the offline success monitoring mechanism we humans are equipped with; our conscience, or inner voice or conditioning or whatever term you like to call it.

The job of this offline success monitoring mechanism is to use past learnt personal and cultural history to guide our actions and choices. In effect we store both our individual histories and cultural histories, our understanding of the rules that arise from these histories and posit trial solutions to the problems we face in everyday life. We then wonder how the environment would react to it based on its history of earlier reactions. To enable this look ahead and guessing, every real world success and failure, personal and cultural, goes into this archive.

The learning process acts in two directions, one way to guess a solution, and another way to guess the environments response. Improved solutions are verified and checkmated against past personal and cultural history before they are enacted in the real world. The feedback of such an act is again fed back into the historical archive for reprocessing and correction; this is a continuous process that keeps our mind busy. Notice that such an offline success monitoring mechanism is only possible for systems that have the luxury of online minds; that means that only we humans and perhaps some other predators may have the luxury of conscience and inner voices.

An offline success monitoring mechanism can and does exist in us and it is logical that its presence and availability is equivalent to the presence of the offline processing mechanisms. The offline learning mechanism uses the offline success monitoring mechanisms to generate and improve offline proactive solutions. Man's mix of mindlessness and mindfulness seem to ensure the almost permanent presence of such a proactive offline mechanism. Man therefore shuttles between the demands of instinctual action and considered conscious action.

While instinctual patterns arise from his evolutionary heritage, the offline learning is newer and current and arises from cultural learning and the offline learning processes and its inbuilt success monitoring mechanisms. Here is where we can see conscious action score over instinct, here is where we set long term strategy that overrules the environmentally correct short term instinctual response, and here is where cultural training can offset natural behavior.

Such simulated offline proactive behavior, whose basic idea seems simple to us, could prove difficult for the processing system. For offline proactive behavior to happen it should make the mind not a theater where vanilla processing happens, but a battleground of options and hypotheses. In this offline learning process with its own success monitoring mechanism, thoughts can loop incessantly looking for a solution.

A good simile is that of a stage performance or a movie, a story runs on the stage, however this is a stage where the director, scriptwriter and actor are one and the same, the entity himself, performing before an imaginary audience that consists of an environment and other people and possibly the entity itself, it is on this stage that hypotheses are tried and as Popper said, die in our stead, the scriptwriter has to find a way out of the conundrums he himself has created or undergone. This calls not only an awareness of the environment but its past history too, not only history, but an interpolation of that history into the present future, it calls for recollecting a chain of events and rechaining it as the entity wishes to, it can become a big job. Developing an AI system like this would be a difficult affair, but then who said that it is easy to be human? We will expand the discussion on this in a later paper.

However notice that even the above discussion helps better our picture of our mind and our sense of I. In our earlier description the concept of I looked like a rather forlorn object necessitated by offline learning demands. Now we see that the "I" and the mind are better fleshed out, these are not just simple shadow entities and shadow environments, these are full fledged learning systems, existing and running within the base learning system, like say Windows running within Linux or vice versa.

We can see that these offline learning systems are actually equipped with all the appurtenances of a basic consciousness based learning system, a sensory boundary image, a learning mechanism, a success monitoring mechanism, archival systems, predictive and proactive behavior, time constraint switches and so on. This is practically an identical mirroring of the basic consciousness based learning mechanism. Its only limitation is that it requires and runs on the basic mechanism and cannot exist without it. Now the explanation for our mind and our sense of I becomes almost realistic, we do recognize this beast as our own.

We see that Descartian Theater to use a Dennet (3) phrase needs to exist, and not only in man, however we need to find the way such a mechanism is actually implemented. The mind is the (offline learning) theater of the brain. The "I" is the online identifier of the mind. This I consumes much of his time, during his wakefulness and even during his sleep, (during sleep he recognizes it only when he dreams). We see why thoughts need to loop incessantly in our brain, more difficult the problem, more unrelenting the swirl. There will always be problems that never go away and people who never give up, so the looping can be infinite and vicious.

Our discussion of the possible origins and nature of nature's learning mechanisms are drawing to a close. We have however barely scratched the surface, what has emerged is a possible substrate mechanism that can display behavior that is far more complex and richly delineated depending on its extent and use. In our effort to provide a basic framework, we have stripped it down so that its rich details do not obscure its base design. The presence of the gaps is obvious in this discussion and will be keenly felt by the discerning reader and will initially lead to much nay saying and shaking of the head. We do have to deal with the proactive mechanism in more detail, but that has to wait.

Notice that we have no quarrels with natural scientists or philosophers, we have only wondered about the base design of consciousness and consciousness enabled learning systems from an engineering point of view. We do not and have not discussed and speculated on the actual implementation of consciousness based learning at the neuronal or other levels. The reader can however see that that we have relied on no external factors and characteristics; the discussion rests on few simple assumptions, is fairly straightforward, and is easily falsifiable.

This hypothesis about the design choices of natural learning systems will make sense only if it is read in conjunction with the theory of evolution. Like for evolution, the reader perforce has to consider evolutionary timescales, pressures and influences, a panoptic view is first necessary to understand the import of the hypothesis before one can map it to the natural world. Any system natural or artificial, experience tells us, rests on a bed of logical foundations. Life and consciousness cannot be any different. This discussion has been an effort to put a logical systemic face to the problem.

Natural scientists can use the perspective to probe and explain some of the mechanisms that make up man's mind and consciousness. If the consciousness and learning systems of man and animal derive from the same roots, then it becomes easier to probe. Our knowledge of natural intelligence systems is still in its early stages and evidence is quickly emerging from various areas of natural science. The real measure of the hypotheses would be the degree of fit to present and emerging evidence. The author would be glad to have and answer relevant feedback on explicit evidence standouts and conflicting opinions.

**Database, Learning, Patterns, Search - A Question of Scale**
Having set the consciousness and learning mechanisms in place, one needs to wonder what kind of archive would satisfy the mechanism and what kind of learning mechanism should be implemented. The author refrains from a detailed discussion of learning, search, pattern identification, archive management, and database processes here.

However, from our discussion we know that time-constraints deny the possibilities of complete processing or complete learning, most of the time. The same applies to search and cognition processes too. On the other hand, there are occasions when complete learning is possible and our offline learning processes do contribute to increasing such a possibility. Most organisms would need to be satisfied with a mix of online and offline processes. These call for an ability on the part of the learning mechanisms to vary learning/search/cognition/response processes based on the available time and environmental urgencies.

This calls for a scaled learning approach, visualize a scaling slider that moves up and down based on the capricious time and response demands of the environment. A scaled learning process can work when the underlying archive is also scaled. Such an archive can also lend itself to scaled cognition and search processes. Let us first visualize a scaled database arrangement and then posit learning, search and pattern identification processes on it.

To visualize this design, imagine a range of hills and a helicopter able to hover over and within it. The data arrangement is like this. Each hill is devoted to a context; it consists of an object and links to other objects. The height and spread of the data and contexts are dependent on contextual or data strength and interlink strength. The top of the hill is populated with high strength data points. This area is generally detail sparse and possibly interlink sparse. In the mid area, the details emerge, and the links multiply and are perhaps stronger. The base is a thicket of low strength relationships among low strength data signals. Therefore data scaling seems to be on data density and relationship density.

# Conscious Intelligent Systems

Natural Intelligence and Consciousness – A Learning System Perspective

## Part I: I X I

Under extreme time constraints, the helicopter can just hover over the hill tops, get an approximate data range and process it for learning or search. Such learning is little more than a random guess, as we go down the hill; it will become data rich and can become an approximation. The learning system can also generate theories based on a simple sampling of the data and then search back the archive for supporting data; we can call this anticipation (**def:** to guess with desire or intent) This allows for solution seeding, which speeds up the learning process.

Natural systems do this guess and test most of the time. When time is available or when iterative learning demands arise, then the helicopter can come down the ranges and exhibit a better quality of learning/search. During deep offline learning, the quality of learning will become better if the time available switch is thrown off, reducing response pressure.

Such scaling of the database and learning processes also predict scaled searches, which is a prime requirement for any entity in time-constrained environments. Scaled searches also imply that the search process can branch off at certain scale intervals; then depending on the response of the entity to the present output, search terms can be modified on the fly. This means the process of search morphs into a tree traversal search, search can expose an object and its link, and search can proceed along probable links. Such a search path can give rise to heuristic search rather than linear search. The search process can also become stop and go, going underground when interrupted by online processing and coming up when online demands weaken.

Scaling helps the pattern identification process also, rather than seek patterns in large masses of data, it can look for patterns at the sparse top end of the data hill and seek to confirm patterns as it traverses down hill. This design can help generate time-delimited responses to learning and search.

Times constrained learning, interrupted or stop and go learning, sub optimal learning, optimal learning, iterative learning, plain guessing or anticipation, quick search, normal search, deep search, tree traversal search, simple pattern discovery, pattern confirmation and pattern linking are all possible with such a database and learning system design. Such scaled learning can sometimes result in data discontinuity to the processing systems and data bridging may be necessary, it may be required to guess a missing link. This calls for fuzzification; the positive side of this activity is insight discovery.

The author wonders if such a scaled learning and database design is possible or if it already exists in the literature. Rather than a master design, it is obvious that there can arise a simple reusable unit like design that can incorporate these features, if there is one, the author has not managed to figure out such things to a degree of confidence as to present it.

**Designs of Nature**

This section was originally part of our basics section. The ideas here form natural extensions of our earlier arguments on nature and consciousness, there might be some repetition. Time constrained readers may skip this and jump directly to the section on understanding.

Our earlier discussion helped us strain the major points and helped us stake out the possible design for a consciousness mechanism; here we see some of the implications of our arguments and the design it begot. Consciousness seems to be life's tool to sustain itself; we do not venture to seek the purpose or origin of life; however we did presume that it acts in a certain manner. We can see that life uses consciousness and intelligence as mechanisms to protect its exhibits and sustain them.

The sum and substance of our consciousness argument is that (given life and its desire to sustain itself) the more stable the environment, the more complex the systems that evolve, then the more difficult, more costly it becomes to build, repair and replace, therefore the more sustenance and protection it demands, and so more the requirement for consciousness. The level of consciousness is proportional to system complexity and sustenance/protection costs.

# Conscious Intelligent Systems

Natural Intelligence and Consciousness – A Learning System Perspective

## Part I: I X I

The advantage of consciousness based rules is that very simple factors drive it, the demand for protection and sustenance, there is no other logic or ideal to which these systems aspire to, system structures and behaviors are oriented towards these demands, in a sense, these systems are the ultimate realists. No wonder then that life has survived in spite of extreme pressures and multiple extinctions; we can judge that it would even survive multiple nuclear winters if niches can ensure that at least some form of life survives.

We know from geological and biological records that extreme environmental variations have always meant extinctions. Environmental variance is the greatest challenge to all life, when we say environment we mean both the animate and inanimate environments and one that includes all inhabitants and their effects on the ecosystem. Why should entire species and genera be wiped out in one single stroke of the evolutionary pen?

From our perspective we see that when environments change they generate sudden learning loads and when natural intelligence systems cannot keep pace with learning loads, then extinction can result. This can also happen to entities that have little morphological disabilities in new environments. We also wonder if excessive learning loads dampen fertility and thus help trigger species extinction.

The interaction of consciousness based directed learning systems and natural environments could throw up a variety of entities with varying levels of consciousness and intelligence. Since environments themselves seem to arise from a complex interaction of forces with no pre-defined trajectory, and consciousness based systems try to dovetail to them, we can see that the trajectory of intelligence system growth paths cannot be defined clearly. What is however clear is that there no ideal design or growth path; all paths are dependent on a mix of historical and current system pressures. We cannot really foresee what kind of organisms and systems may emerge in the process.

Therefore setting the clock back 65 million years and rerunning the earth's evolutionary program does not guarantee the rise of humans, it could be something different too. The prospect of human like alien forms and ruminations on their intellectual capacities also comes under the same hammer; we simply cannot say. Here we see that human mental involution itself arises out of a certain combination of enablers and constraints, without such factors humans would never have emerged. The most intelligent entity on this entity would have been an ape for reasons it wouldn't even know or get to.

One thus sees a parallel with what Darwin wanted the world to see; that there is no master design in Nature, evolution is an intersection of multiple forces; here we see that learning system abilities and learning system paths form important factors in evolution. In nature, such conditions can give rise to a confusing multiplicity of platforms and levels. Darwin's first success was to prove that all this seeming multiplicity of life follows a logical inheritance path, but with no logic of its own. (Darwin's idea is logical and therefore extremely seductive to us as learning systems, which are always under pressure to seek answers to close out existing questions. As to the actual evidence and its interpretation, there seems to be a fair bit of debate even among evolutionists). Having figured out the path of inheritance, Darwin did wonder as to the logic of inheritance and the resultant divergence. How did successors so divergent and varied emerge from their simple ancestors?

"But at that time I overlooked one problem of great importance; and it is astonishing to me, except on the principle of Columbus and his egg, how I could have overlooked it and its solution. This problem is the tendency in organic beings descended from the same stock to diverge in character as they become modified. That they have diverged greatly is obvious from the manner in which species of all kinds can be classed under genera, genera under families, families under sub-orders and so forth; and I can remember the very spot in the road, whilst in my carriage, when to my joy the solution occurred to me; and this was long after I had come to Down. The solution, as I believe, is that the modified offspring of all dominant and increasing forms tend to become adapted to many and highly diversified places in the economy of nature." - Excerpted From Darwin's Biography – Source: Project Gutenberg

# Conscious Intelligent Systems
Natural Intelligence and Consciousness – A Learning System Perspective
## Part I: I X I

He concludes that the process of speciation can arise out of the internal pressure of entities tending to dovetail themselves to their local environments. From our learning system perspective we can see how such a solution can arise from the activity of directed learning, but that still begs the question as to how these solutions find their way back into the genes. There are as many unanswered questions on this subject as there are questioned answers. Let us not add our bit to it. From a simpler straightforward perspective, one can easily see that speciation can arise on two counts, from generations of isolation within environments and from generations of isolation within food chains. Probably the former corresponds to macroevolution and the latter to microevolution. There could be more reasons we are not presently aware of.

From our perspective, we can see that learning archives can vary both on an individual basis as each entity responds to its peculiar view of the environment, but also between generations, as they tend to dovetail to their current environments. Our current state of knowledge posits gene passing as the way of reproduction and therefore learning retention, therefore from our perspective we should expect to see some genetic write back with each individual and generation as it learns to interact successfully with its environment. Variation in lesson passing is to be expected not only with each individual but also with each generation, though they could be so small as to be unnoticeable.

It is understandable that all learning need not be reflected back into the pattern passing mechanisms, only those with a certain level of persistence may need to, one can posit a write back threshold, however one sees that the entity or its species cannot get away with zero write back in the long term. Write back requirements can also reduce with the presence of the cultural learning mechanism, which reduces the need to write back minor environmental variations to the database, however this will mean that the loss of incubation translates to a steeper learning curve for the entity.

We have conjectured that all natural intelligence (directed learning systems) arises as a result of the presence of consciousness and the desire of life to sustain itself. One can foresee that consciousness enabled learning systems will use the intentional, directed learning mechanism to adapt closely to environments, probably until they hit morphological limits. The next natural question is that of how do morphologies change. Darwin's answer was evolution by natural selection.

On first inspection, natural selection looks like a poor choice, long geographical periods, multiple generations, multiple lifetimes, more evolutionary failures than successes, and too damn resource intensive. Despite the rather discomforting observation that evolutionists have still not explained the evolution of the rather long neck of the giraffe convincingly, either from fossil based evidence or otherwise, the fact seems to be that there is no other logical process that can successfully account for the millions of species that inherit the earth and have passed through the earth till date. No alternate theories have risen to successfully challenge Darwin's basic theory. The theory of evolution does look a little like Charlie Chaplin, all askew and in ill-fitting attire, but then creationism, which is more an argument and less a theory looks more like the king that wore no clothes.

From our perspective, we see natural selection as an environmental success-monitoring mechanism that enables a directed growth of living systems. We can see that paths and results that rise out of success monitoring mechanisms could be inherently messy. Such results do not show or follow a clear linear hierarchy; they are more the sum of parts. A theory like the theory of evolution offers us a logical window that allows us to discern the broad logical stream that underlies the seeming chaos.

How and why did complex species arise at all? The simple and most natural answer is that because environments for some reason stayed stable and benevolent. The other answer is that it is rare for extinctions to completely destroy the entire population; the environment has no such aim, therefore survivors did manage to survive in ecological and environmental niches. For them there is no going backward, the only way is forward or sideward. When environments revert back to stability, it is time for whatever is left of life to pick up where it stopped. New interconnections may also arise between the survivors in place of older links that were lost during the unstable phase, and these interconnections may themselves induce modifications.

When we marry our consciousness driven, archive-based, pattern-based learning systems (with their inherent tendency for environment comfort) with environments that can show variance, we can see that processes like natural selection and evolution can and will naturally arise out of such a marriage. We see that pattern-rewriting difficulty can complement or perhaps drives Darwinian selection, pushing the extinction envelope further and faster. We recognize that Darwinian natural selection scenarios can rise not only from morphological disabilities in the new environment but also from learning system problems.

It is obvious that there is a clear call for a mechanism that enables easy relearning and rewriting of patterns. We can speculate that such architecture arose, not from the ashes of patterns but astride the pattern writing architecture, we see cultural writing as Nature's solution in the face of so much extinction as a way out. The rise of what we call cultural writing; organisms acquiring learning in their lifetime and passing it on to their descendants, bypassing the slower instinctual learning mechanism, perhaps helped slow down the rate of species turnover and perhaps did help stop the high end of Darwinian evolution in its tracks, culminating in man, where the cultural rewriting process can be considered to be well entrenched.

With cultural writing, learning, relearning and rewriting becomes easier. Depending on the efficacy of its directed learning mechanism, the entity could relearn and apply the lessons of life to itself and teach its descendants. Therefore as long as a species does not face extreme morphological constraints and demands or resource scarcity, it has a better chance of keeping track of its environment, which in turn increases the chances of survival. Cultural learning offers a waiting buffer facility for learning transference, only when environmental conditions turn persistent, is it required to transfer learning to the instinctual pattern based mechanisms.

Learning transfer does bring us to reproduction. Asexual reproduction makes sense even from our perspective. A learning system bequeaths its learning to its inheritors to carry on, in the interests of life's sustenance. Notice that in an ideal environment with multiple inhabitants, the directed learning process tends to induce individual divergence over a common base; variation is the norm, except for twins in highly similar environments. Higher the intelligence and more varied the environment and more varied the environment worldview, higher the divergence and individual variation would be. The net benefit of social grouping must therefore be higher for groups to arise. The roots of sexual reproduction seem to be an extreme version of such grouping instincts, a sense of formalized give and take, of learning or genes. We certainly need to know more about the connection between the genes and learning for us to even speculate further on this point.

Life, from a panoptic point of view seems to have followed two major paths in system development. In the static path, like trees and other flora, it took system replacement as easier than system protection, thus favoring a system that could easily replicate and conquer the world.

Static entities have few sensor requirements simply because the scope for system protection is low, we can perhaps ask rhetorically how it would help a tree to have a pair of eyes. These lower sensory requirements directly translate to lesser levels of consciousness and learning. For trees and similar static entities consciousness consists of a distributed but interconnected network that can realign itself quickly in case of system damage.

On the sustenance angle too, static systems tend to be simple systems that learn to survive on what is available. It could be argued that trees develop patterns and develop proactive behavior, but the extent and richness of such behavior is lower when compared to such behavior by their mobile counterparts. Plants and trees are evidence that an alternate to centralized consciousness exists within the bounds of natural intelligent system design.

Consciousness demands on the mobile side are the other way around of stasis; components increasingly become complex, sensor wiring gets dense, input and output centralization becomes high and replacement gets so costly that consciousness and intelligence has evolved to sustain and protect it. As regards food, for mobile entities, to forage is life. Life is also exceptionally risky.

# Conscious Intelligent Systems

Natural Intelligence and Consciousness – A Learning System Perspective

## Part I: I X I

These contribute to higher consciousness levels and higher intelligence for mobile entities. Mobile entities also need good environmental awareness.

As we move up the mobile hierarchy, the patterns multiply and environmental learning demand again determines their learning ability. Learning ability is not something that entities can acquire overnight; it is a complex intersection of environmental challenges and demands, their evolutionary history, their existing level of consciousness, the available rest time, and internal system demands. The environment also determines where and how these learning abilities are used. We did see that the dolphin and man may have similar intelligence levels but man's environment put him on top and made the dolphins his toys, it could well have been otherwise.

In our basics section, we did discuss that learning from scratch is not only impossible but also improbable for a sufficiently advanced natural learning system. Even given the facility of incubation, such an ab initio process will take a long, long time. Nature's incubation requirements for its entities also seem to vary on a mix of pattern complexity and environment demands, some come OS ready, ready to perform, some come with pattern frameworks, with data filled in during incubation time, parental or otherwise.

We see that the amount of learning that any entity or species can do within its lifetime is constrained by its processing abilities and other external factors. It looks logical to propose that longer lifetimes would mean more learning. Why did not Nature take such an approach? It looks like there are environmental and morphological constraints to such an approach.

The availability of food resources, the cyclic nature of the weather and climate, the wear and tear induced by environmental demands all constrain the idea of long lifetimes. Nature seems to face a dilemma in choosing between learning ability and such natural constraints. The fact that an environment may itself show variation and thus a sudden spurt in learning load, within a single lifetime is also a perennial worry, because the available learning ability may not suffice to take care of such variation. Shorter lifetimes allow parceling of the problem into manageable parts.

The author presumes that nature's choice of rapid life turnover and pattern passing through reproduction emerged as a golden mean solution to such a dilemma. The lifetime of a natural entity is therefore a probable intersection of its learning abilities and such ecological and environmental constraints. In environments of stability and minimal wear and tear, it is possible that learning ability may be a restriction to longer lifetimes. He does not know if the evidence will allow for a supposition that the lifetimes of natural organisms are not only a reflection of their maintenance costs, but are also a reflection of their learning abilities. Full Stop!

**Post Script**: As regards the application of the mind mechanisms to nature, our discussions have barely scratched the surface. We have sacrificed much of the details that are critical to natural science to gain the freedom to say that this is perhaps how it all ought to be. This is at best a perilous approach towards the natural sciences, as even evolutionists are finding, there will always be standouts that need of explanations within a broad logical framework. Making a law that explains natural living systems is more difficult that making physical laws. No past or future biologist in the past and future can afford Newton's freedom when he asserted that he did not speculate. Nature's rules do not seem to follow the rigidity of physical law. As regards our proposed mechanism, it is for the biological and associated sciences to accept/reject/extend it or to fill in the gaps with detail. We have merely staked out the possible substrate design.

**References:**
1. Turing, A.M (1950), Computing Machinery and Intelligence Mind 49: 433-460.
2. Darwin, Charles (1871), The Descent of Man
3. Dennett, D.C. & Kinsbourne, M (1995) Time and the observer-Behavioral and Brain Sciences 15 (2): 183-247
4. Joel Achenbach (2004) Who's Driving? - National Geographic, Nov 2004
5. Gayathree U, (2006) Conscious Intelligent Systems Part II - Mind, Thought, Language, and Understanding,